\documentclass[lettersize,journal]{IEEEtran}
\usepackage{amsmath,amsfonts}
\usepackage{algorithm}
\usepackage{array}
\usepackage[caption=false,font=normalsize,labelfont=sf,textfont=sf]{subfig}
\usepackage{textcomp}
\usepackage{stfloats}
\usepackage{url}
\usepackage{verbatim}
\usepackage{graphicx}
\usepackage{cite}
\usepackage{booktabs}
\usepackage{multirow}
\usepackage{diagbox}
\usepackage{subfig}
\usepackage{epstopdf}
\usepackage{CJK}
\usepackage{multirow}
\usepackage{amsmath}
\usepackage{color}
\usepackage[T1]{fontenc}
\usepackage{array}
\usepackage{verbatim}
\usepackage{multirow}
\usepackage[dvipsnames]{xcolor}
\usepackage{pdfpages}
\usepackage{colortbl}
\usepackage{tcolorbox}
\usepackage{ulem}
\usepackage{pifont}
\usepackage{algpseudocode}

\renewcommand{\algorithmicensure}

\usepackage{hyperref}

\def\BibTeX{{\rm B\kern-.05em{\sc i\kern-.025em b}\kern-.08em
T\kern-.1667em\lower.7ex\hbox{E}\kern-.125emX}}

\begin{document}

\color{black}
\title{Copy-Move Forgery Detection and Question Answering for Remote Sensing Image}
\color{black}
\author{IEEE Publication Technology,~\IEEEmembership{Staff,~IEEE,}
}
\author{Ze Zhang\textsuperscript{\dag},  Enyuan Zhao\textsuperscript{\dag}, Di Niu, Jie Nie\textsuperscript{*}, \emph{Member, IEEE}, \\Xinyue Liang,  \emph{Member, IEEE}, Lei Huang, \emph{Member, IEEE} 

\thanks{This work was supported by the National Natural Science Foundation of China (U22A200536 and U23A20320), and the Research and Innovation Capacity Support Program for Young Faculty of the Ministry of Education of China (SRICSPYF-BS2025003).}

\thanks{Ze Zhang, Di Niu, Xinyue Liang and Lei Huang are with the Faculty of Information Science and Engineering, Ocean University of China, Qingdao, 266100, China.

Enyuan Zhao is with the Hangzhou Institute for Advanced Study, University of Chinese Academy of Sciences, Hangzhou, 310024, China.

Jie Nie is with the Faculty of Information Science and Engineering, Ocean University of China, Qingdao, 266100, China, State Key Laboratory of Physical Oceanography, Ocean University of China, Qingdao 266100, China and Shandong Key Laboratory of Intelligent Sensing Chips and System, Ocean University of China, Qingdao 266000, China.

{\dag}: These authors have contributed equally to this work and share first authorship.

\emph{Jie Nie is the corresponding author; Emails: niejie@ouc.edu.cn.}
}
}


\IEEEpubid{}

\maketitle

\begin{abstract}
Driven by practical demands in land resource monitoring and national defense security, this paper introduces the Remote Sensing Copy-Move Question Answering (RSCMQA) task. Unlike traditional Remote Sensing Visual Question Answering (RSVQA), RSCMQA focuses on interpreting complex tampering scenarios and inferring relationships between objects. We present a suite of global RSCMQA datasets, comprising images from 29 different regions across 14 countries. Specifically, we propose five distinct datasets, including the basic dataset RS-CMQA, the category-balanced dataset RS-CMQA-B, the high-authenticity dataset Real-RSCM, the extended dataset RS-TQA, and the extended category-balanced dataset RS-TQA-B. These datasets fill a critical gap in the field while ensuring comprehensiveness, balance, and challenging scenarios. Furthermore, we introduce a region-discrimination-guided multimodal copy-move forgery perception framework (CMFPF), which enhances the accuracy of answering questions about tampered images by leveraging prompts about the differences and connections between the source and tampered regions. Extensive experiments demonstrate that our method establishes a stronger benchmark for RSCMQA compared to general VQA and RSVQA models. Our datasets and code are publicly available at \href{https://github.com/shenyedepisa/RSCMQA}{https://github.com/shenyedepisa/RSCMQA}. 
\end{abstract}


\section{Introduction}
\color{black}

\IEEEPARstart{H}igh-resolution remote sensing images are instrumental in rapidly acquiring critical information \cite{Zvonkov2023OpenMapFlow, huang2025survey, lenton2024remotely}. These images can be utilized for soil moisture inversion, monitoring forest coverage, enhancing ecological protection policies, and integrating multi-source remote sensing data to depict urban development trends and strengthen urban management \cite{wang2024skyscript}. Additionally, extracting valuable information from remote sensing images is crucial for national defense security monitoring, especially for situational awareness during wartime. However, the content of digital remote sensing images is susceptible to manipulation or forgery, which can be achieved by copying objects from the original image to another location. Copy-move image forgery involves copying a specific region of the image (source region) to another location within the same image (tampered region). Since the tampered and source regions originate from the same image, their optical characteristics are nearly identical, which significantly increasing the difficulty of detecting the tampered areas.

Detecting tampering in remote sensing images holds significant academic and practical value. Traditional methods for copy-move forgery detection (CMFD) in natural images primarily encompass block-based, keypoint-based, and deep learning-based approaches \cite{christlein2012evaluation,abd2016copy,elaskily2023survey}. However, the unique perspectives inherent in remote sensing images, coupled with extensive monitoring areas, numerous small objects, and limited resolution, exacerbate the challenges faced by general CMFD methods in accurately identifying tampered regions. Specifically, extracting high-level semantic information from the source and tampered regions in complex tampering scenarios is challenging, which further impedes researchers' ability to access and interpret these critical tampering details.

\begin{figure}[!t]
    \centering
    \includegraphics[width=1\linewidth]{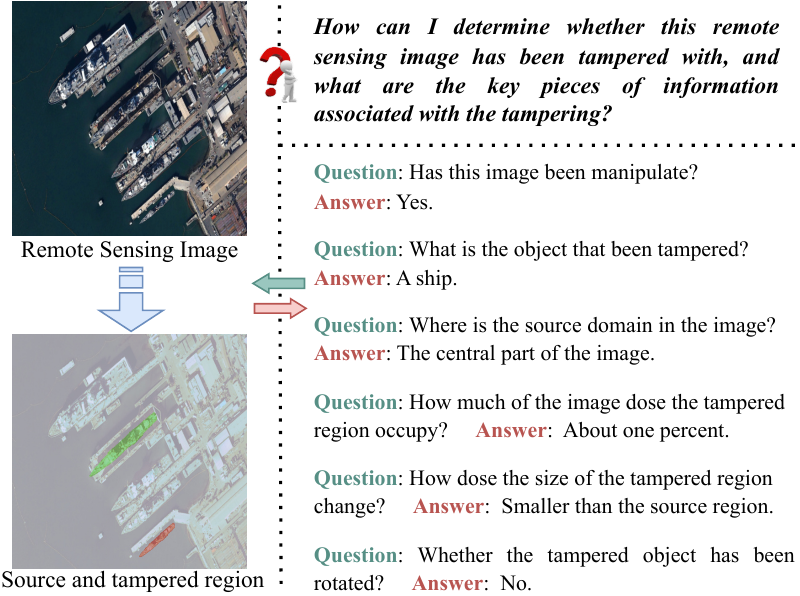}
    \caption{Example of using question-answering method to obtain key information about remote sensing image tampering.
    }
    \label{fig:motivation}
\end{figure}

Remote Sensing Visual Question Answering (RSVQA) leverages neural networks, driven by textual inputs, to enable the perception of remote sensing images, thereby surmounting the efficiency constraints of information extraction for remote sensing interpretation tasks. Preliminary VQA methods and datasets specific to the remote sensing domain, introduced by scholars such as Lobry\cite{lobry2020rsvqa}, Zheng\cite{Zheng2021Mutual}, and Yuan\cite{yuan2022easy}, have established the foundational framework for the RSVQA task. Building upon existing research, the question-and-answer framework is identified as a feasible and effective approach for accurately and efficiently extracting tampering-related information from remote sensing images, as illustrated in Figure \ref{fig:motivation}. Nonetheless, in light of the practical demands of national defense security and land resource monitoring, current RSVQA methodologies fall short in their capacity to accurately extract high-level attributes, such as source and tampered regions in copy-move tampering scenarios. Compared with existing RSVQA and CMFD tasks, RSCMQA exhibits significant differences in research objectives and application scenarios. RSVQA mainly focuses on semantic understanding of untampered imagery and cannot address questions related to image authenticity; classical CMFD methods concentrate on forgery detection in natural images, typically producing binary masks or region localization without multimodal reasoning. In contrast, RSCMQA is the first to integrate tampering detection with question answering, enabling a transition from “detection” to “understanding” in remote sensing tampering scenarios. This research gap directly motivates the three challenges addressed in this work.

1) Neglect of Copy-Move Forgery Research: Current research on RSVQA primarily focuses on extracting information from untampered remote sensing images, emphasizing basic geographic data to address general questions. However, existing approaches lack a dedicated question-answering system designed to handle the complexities introduced by image tampering in remote sensing scenarios.

2) Lack of Comprehensive and Balanced Datasets: The RSVQA dataset, encompassing image-level, semantic-level, and finer-grained questions, suffers from a highly imbalanced distribution of question types, potentially introducing biases into the question-answering models. Low-quality datasets diminish the task’s significance and challenge, reducing its practical value.

3) Challenges in Perceiving Tampered Images: Tampered remote sensing images present significant challenges to the model’s discriminative capabilities, particularly in accurately discerning the attributes and spatial relationships of source and tampered regions.

To this end, five RSCMQA datasets are introduced, along with a region-discrimination-guided multimodal copy-move forgery perception framework (CMFPF), designed to advance the complex RSCMQA task. The principal contributions of this work are as follows:
\begin{itemize}
    \item We introduce five datasets tailored for the RSCMQA task, with raw data collected from over 29 distinct regions across more than 14 countries. RS-CMQA comprises 118k images and 1.37 million CM-Q-A triplets. To mitigate category bias, weighted random sampling is applied to the RS-CMQA dataset, yielding a balanced subset, RS-CMQA-B, where B denotes ``balance.'' Additionally, we present Real-RSCM, a manually annotated high-quality dataset featuring tampering instances that are subtle, realistic, and logically coherent. Furthermore, RS-TQA extends RS-CMQA by incorporating blurred tampered images, accompanied by its balanced counterpart, RS-TQA-B. Collectively, RS-CMQA establishes a foundational benchmark for RSCMQA, while RS-CMQA-B addresses long-tail data distribution and category bias. Real-RSCM enhances realism, posing greater challenges, and RS-TQA/RS-TQA-B introduce blurred tampering to assess model generalization. These datasets bridge a critical gap, ensuring comprehensiveness, balance, challenge, and generalization, thereby providing a rigorous benchmark for evaluating RSCMQA models. 

    \item We propose the Region-Discrimination-Guided Multimodal Copy-Move Forgery Perception Framework (CMFPF), which integrates pixel-level visual prompts and textual prompts to guide multimodal reasoning. A key methodological innovation is the Source-Tampering Mutual Aggregation (STMA) module. STMA obtains reconstruction masks for the source and tampered regions from the forgery detection network and processes them through two independent encoders, producing complementary feature representations: one set highlights differences between source and tampered regions, while the other captures their correlations and similarities. By explicitly modeling both differences and similarities, STMA injects forgery-aware cues into the textual modality via Cross-Attention, thereby bridging low-level tampering detection with high-level question answering. This dual-channel design ensures that the language branch is contextually guided by visual evidence of tampering, enabling fine-grained reasoning about semantic and spatial relations.
    
    \item Through extensive experiments on the proposed datasets, we establish a strong benchmark for the RSCMQA task. Our framework consistently outperforms existing baselines, demonstrating the effectiveness of CMFPF and the necessity of tampering-aware multimodal reasoning. These experimental results not only validate the methodological contributions but also provide a rigorous reference benchmark for future research in tampering-aware remote sensing QA.
\end{itemize}

\section{Related Work}\subsection{Copy-Move Forgery Detection }

Image copy-move forgery involves the manipulation of an image by copying and relocating entities within it. The primary motivations for such manipulations are either to conceal an element within the image or to emphasize a particular object. Traditional copy-move detection algorithms typically rely on stringent prior knowledge of image properties, such as edge sharpness and local features, and are generally classified into block-based methods and keypoint-based methods. Block-based methods, such as Principal Component Analysis (PCA)\cite{5438805}, Discrete Wavelet Transform (DWT)\cite{singh2022copy}, and Fourier Transform (FT)\cite{ketenci2013copy}, require segmenting images into overlapping blocks and processing each block individually, which significantly increases computational costs. Keypoint-based methods, including SIFT\cite{gan2022novel}, SURF\cite{kumari2023image}, TRIANGLE\cite{7123632}, and ORB\cite{kaushal2024orbcmfd}, offer more flexible feature extraction but struggle with smooth regions lacking distinct boundaries.

Given the exponential increase in image data, manually designing priors has become impractical. Consequently, deep learning-based methods now dominate CMFD research. BusterNet\cite{wu2018busternet} introduced a parallel dual-branch neural network for separate detection of source and tampered regions. Chen\cite{chen2020serial} then transitioned to a serial approach to resolve feature consistency issues. Islam\cite{islam2020doa} pioneered the use of Generative Adversarial Networks (GANs) in CMFD, enhancing localization accuracy. Liu\cite{liu2021two} combined keypoint extraction with deep learning to improve forgery localization through feature point matching. CMCF-Net\cite{xiong2023cmcf} uses a stacked fusion model to focus on suspicious objects at different scales. UCM-Net\cite{weng2023ucm} treats copy-move forgery as a semantic segmentation task, employing a multi-scale segmentation network for tampered area identification. Wang\cite{wang2024object} proposed an approach that first estimates similar regions coarsely, followed by object-level matching between source and tampered regions.

Current research on copy-move forgery detection, through both traditional and deep learning methods, largely focuses on object detection, which is insufficient for remote sensing images due to their noisy content. This compromises accuracy and fails to provide sufficient information. Thus, integrating copy-move forgery detection into multimodal question-answering tasks is essential. Additionally, the publicly available datasets that underpin CMFD tasks, such as CoMoFoD\cite{tralic2013comofod}, COVERAGE\cite{wen2016coverage}, MMTDSet\cite{xu2025fakeshield}, and MICC\cite{amerini2013copy}, are primarily designed for natural images. The necessity of establishing specialized datasets has been demonstrated by research on ID\cite{mahfoudi2021cmid} forgery detection and medical image\cite{shao2024detecting} forgery detection. Overall, the creation of a copy-move forgery dataset specific to remote sensing images, along with the design of corresponding question-answering models, represents an urgent research priority.

Although classical CMFD methods, ranging from block-based and keypoint-based approaches to deep learning models, have achieved notable progress in natural image forgery detection, they remain limited in remote sensing scenarios. The complexity of multi-scale objects, noisy backgrounds, and low-resolution imagery significantly reduces their effectiveness. More importantly, CMFD methods typically output binary masks or localized regions, without providing forgery-oriented multimodal reasoning or multimodal integration. In contrast, RSCMQA advances beyond these approaches by embedding tampering-aware cues directly into a question answering framework, enabling not only detection but also interpretation of tampering in remote sensing images. Crucially, RSCMQA extracts key information from both source and tampered regions—such as spatial relations, object attributes, and transformation patterns—and leverages it to support fine-grained multimodal reasoning.

\subsection{Remote Sensing Visual Question Answering}

The RSVQA task enables researchers to query remote sensing images using customized multimodal question-answering techniques, thereby obtaining advanced information specific to image content or spatial dependencies among visible objects. Lobry\cite{lobry2020rsvqa} introduced the initial RSVQA model. Building on this, Bazi\cite{Bazi2022Bi-Modal} incorporated a Transformer-based VQA method. Chappuis\cite{chappuis2022PromptRSVQA} classified image information and generated textual prompts, which were then input into a language model for answer prediction. Yuan\cite{yuan2022easy} proposed a language-guided approach with a soft weighting strategy to direct image attention progressively from easy to hard. Siebert\cite{Siebert2022MultiModal} employed the VisualBERT model\cite{li2019visualbert}  to better learn joint representations. Lucrezia\cite{tosato2024segmentation} and Wang\cite{wang2024earthvqanet} used segmentation masks to guide the model's attention to critical image information. Albanwan\cite{albanwan2024image} analyzed the challenges in remote sensing image change detection and expanded the understanding of remote sensing image fusion. ChangeVQA\cite{yuan2022change} detects regional changes in images captured at the same location over different time periods. While regional change detection in remote sensing images has received attention, current research lacks the extraction of critical information from tampered regions, failing to meet the fine-grained perception needs of the RSCMQA task.

On the other hand, high-quality publicly available datasets that support RSVQA research are relatively scarce. The first RSVQA dataset\cite{lobry2020rsvqa} was introduced in 2020, with QA pairs derived from OSM and images sourced from Sentinel-2 and other sensors. The RSIVQA dataset\cite{Zheng2021Mutual} was automatically generated from existing classification and object detection datasets such as AID\cite{xia2017aid} and HRRSD\cite{zhang2019hierarchical}. The FloodNet dataset\cite{Rahnemoonfar2021Floodnet} was designed for disaster assessment, primarily focusing on the inundation of roads and buildings. EarthVQA\cite{wang2024earthvqa} encompasses various object analysis and comprehensive analysis questions, including spatial or semantic analyses of more than three objects.

\begin{figure}[!t]
\centering
\includegraphics[width=.95\linewidth]{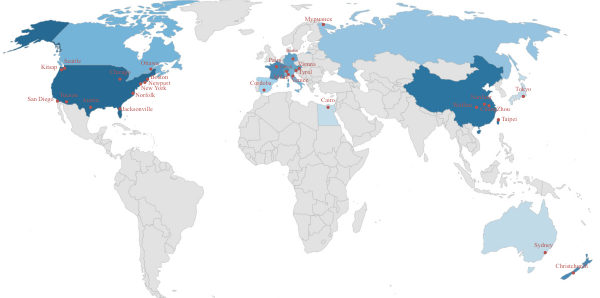}
\caption{Raw images distribution in RS-CMQA dataset.}
\label{fig:fig2}
\end{figure}

These datasets transition from simple questions to complex reasoning, advancing the multimodal remote sensing image community. However, prior studies have not addressed question-answering related to remote sensing image tampering. Additionally, remote sensing QA datasets often suffer from severe data imbalance. The RSVQA-LR dataset\cite{lobry2020rsvqa}, a seminal dataset in this field, exhibits a disparity of over fortyfold between the least and most frequent question categories. Similarly, the latest research, the EarthVQA dataset\cite{wang2024earthvqa}, includes 166 different answers, with the top five answers accounting for 91\% of the total questions. Such severe imbalance may introduce erroneous bias into models and affect the fairness of model evaluation. Therefore, providing tampering-based QA annotations for images, while ensuring both complexity and balance in the dataset, is a crucial focus of dataset development.

\begin{algorithm}[!t]\caption{CM-Q-A triples generation algorithm}
\label{alg:algorithm1}\small
\textbf{Input}: Untampered Images $\boldsymbol{imgs}$, Instance Masks $\boldsymbol{objs}$\\
\textbf{Output}: Tampered Image $\boldsymbol{img}$, Source Region Mask $\boldsymbol{m_s}$, Tampered Region Mask $\boldsymbol{m_t}$, CM-Q-A Triple $\boldsymbol{cmqa}$

\begin{algorithmic}[1] 
\For{$\boldsymbol{img}$ \textbf{in} $\boldsymbol{imgs}$}
\State$\boldsymbol{img}$.manualSelection()
\EndFor
\For{$\boldsymbol{img}$ 
\textbf{in} $\boldsymbol{imgs}$}
\For{$\boldsymbol{obj}$ 
\textbf{in} $\boldsymbol{objs}$}
\If {obj is complete \textbf{and} suitable in size}
\State $\boldsymbol{m_s}$.create($\boldsymbol{obj}$).save()
\State tamper = Choice(CMQA, TQA)
\State $\boldsymbol{obj}$.randomCopy()
\State $\boldsymbol{obj}$.randomRotate()
\State $\boldsymbol{obj}$.randomScale()
\State $\boldsymbol{m_t}$.create($\boldsymbol{obj}$).save()
\State $\boldsymbol{img}$ = copyMove($\boldsymbol{img}$,$\boldsymbol{obj}$)
\For{$\boldsymbol{n}$ \textbf{in} range[1, 15]}
\State $\boldsymbol{cmqa}$.create($\boldsymbol{img}$, $Q_{n}$, $A_{n}$).save()
\EndFor
\State // {for RS-TQA dataset.}
\If{tamper == TQA}
\State $\boldsymbol{m_t}$.create($\boldsymbol{obj}$).save()
\State blur = random.choice(Gaussian, mosaic, daub)
\State $\boldsymbol{img}$ = blur($\boldsymbol{img}$,$\boldsymbol{obj}$)
\For{$\boldsymbol{n}$ 
\textbf{in} [1, 2, 3, 4, 5, 6, 9, 10]}
\State $\boldsymbol{tqa}$.create($\boldsymbol{img}$, $Q_{n}$, $A_{n}$).save()
\EndFor
        \EndIf
        \State$\boldsymbol{img}$.save()
    \EndIf
\EndFor

\EndFor\end{algorithmic}\end{algorithm}

Although recent works such as EarthVQA\cite{wang2024earthvqa} have extended RSVQA to encompass more complex reasoning tasks, including spatial and semantic relations among multiple objects, their applicability to tampering-aware scenarios remains fundamentally constrained. Designed for untampered imagery, EarthVQA cannot address questions of authenticity or manipulation, and its reliance on global segmentation-mask modeling—even with larger architectures—does not meet the core requirement of perceiving source and tampered regions or explicitly modeling their correlations and differences. More critically, the current landscape lacks any dataset specifically tailored for RSCMQA, representing a fundamental gap in supporting tampering-aware reasoning. Beyond this absence, existing RSVQA datasets also suffer from inconsistent quality and pervasive imbalance in question and answer distributions, which introduces bias into model training and evaluation. RSCMQA addresses these limitations by providing tampering-oriented QA annotations, balanced dataset construction, and multimodal reasoning explicitly guided by tampering detection. In doing so, RSCMQA bridges the gap between forgery detection and visual question answering, thereby establishing a novel paradigm that advances beyond both classical RSVQA and CMFD methods.

\section{Dataset construction}

\subsection{Data processing and tampering generation algorithms}

The original images for the RS-CMQA dataset were selected from the LoveDA \cite{wang_2021_loveda}, IAILD \cite{maggiori2017can}, LAISFO \cite{waqas2019isaid}, WHU-Building \cite{ji2018fully}, DroneDeploy\footnote{\url{https://github.com/dronedeploy/dd-ml-segmentation-benchmark}}, HRSC \cite{liu2017high}, and iSAID \cite{kaiser2017learning} datasets. All images were cropped and resized to a resolution of 512 × 512 pixels. After manual screening, we obtained 52,286 high-quality remote sensing images. 
These images originate from at least 29 regions across 14 countries, as shown in Figure \ref{fig:fig2}.

\begin{figure}[!t]
    \centering
    \includegraphics[width=1\linewidth]{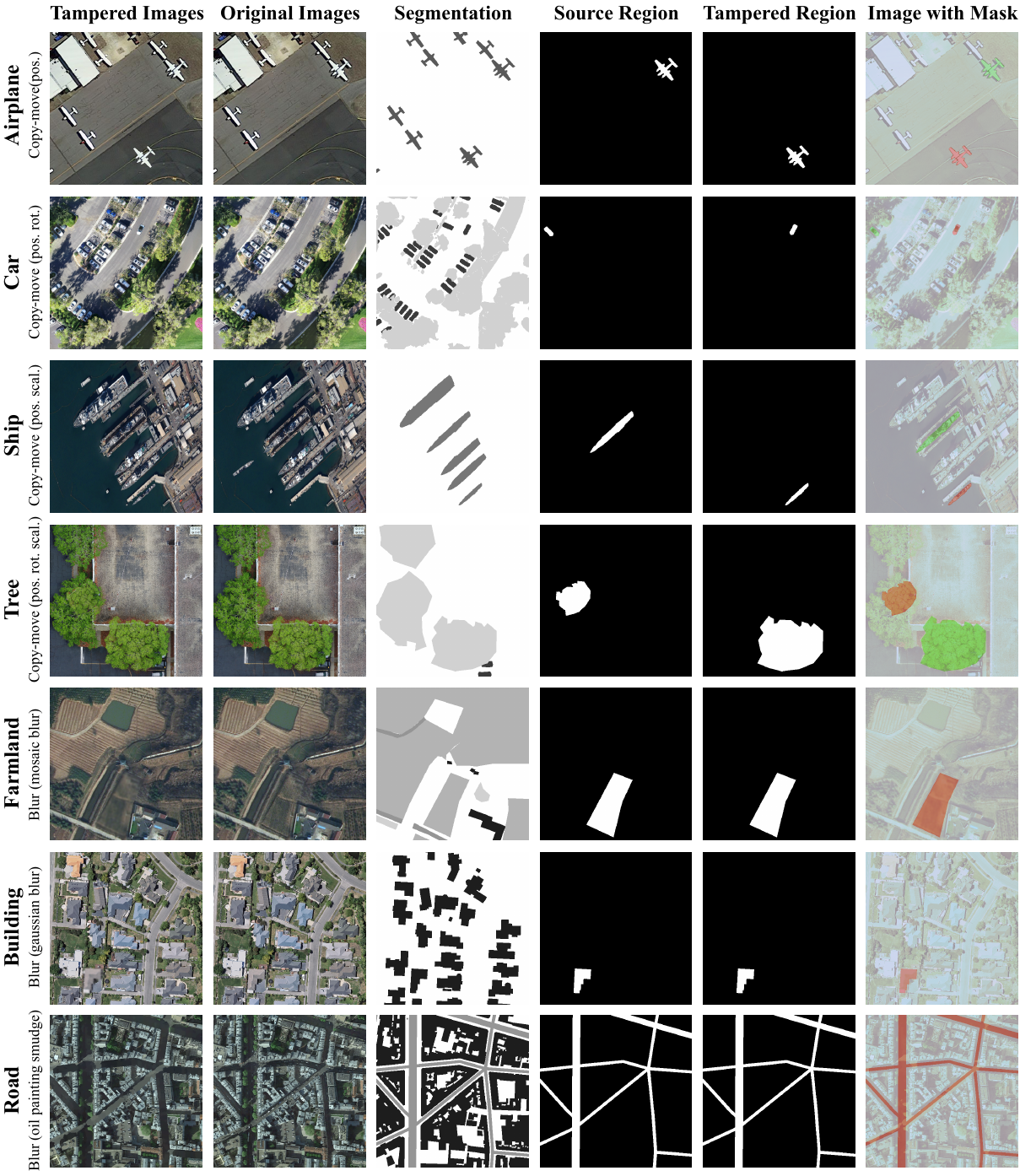}
    \caption{Examples of tampered images, original images, segmentation masks, source region masks, and tampered region masks in the dataset.
    }
    \label{fig:example}
\end{figure}

In this study, we selected seven types of salient targets for tampering: vehicles, airplanes, ships, buildings, roads, trees, and farmland. The chosen tampering targets are independent, separable regions occupying 0.1\%-15\% of the image area, ensuring that all tampered entities, except for roads, are fully presented in the images. 

\begin{figure*}[!t]
    \centering
    \includegraphics[width=\textwidth]{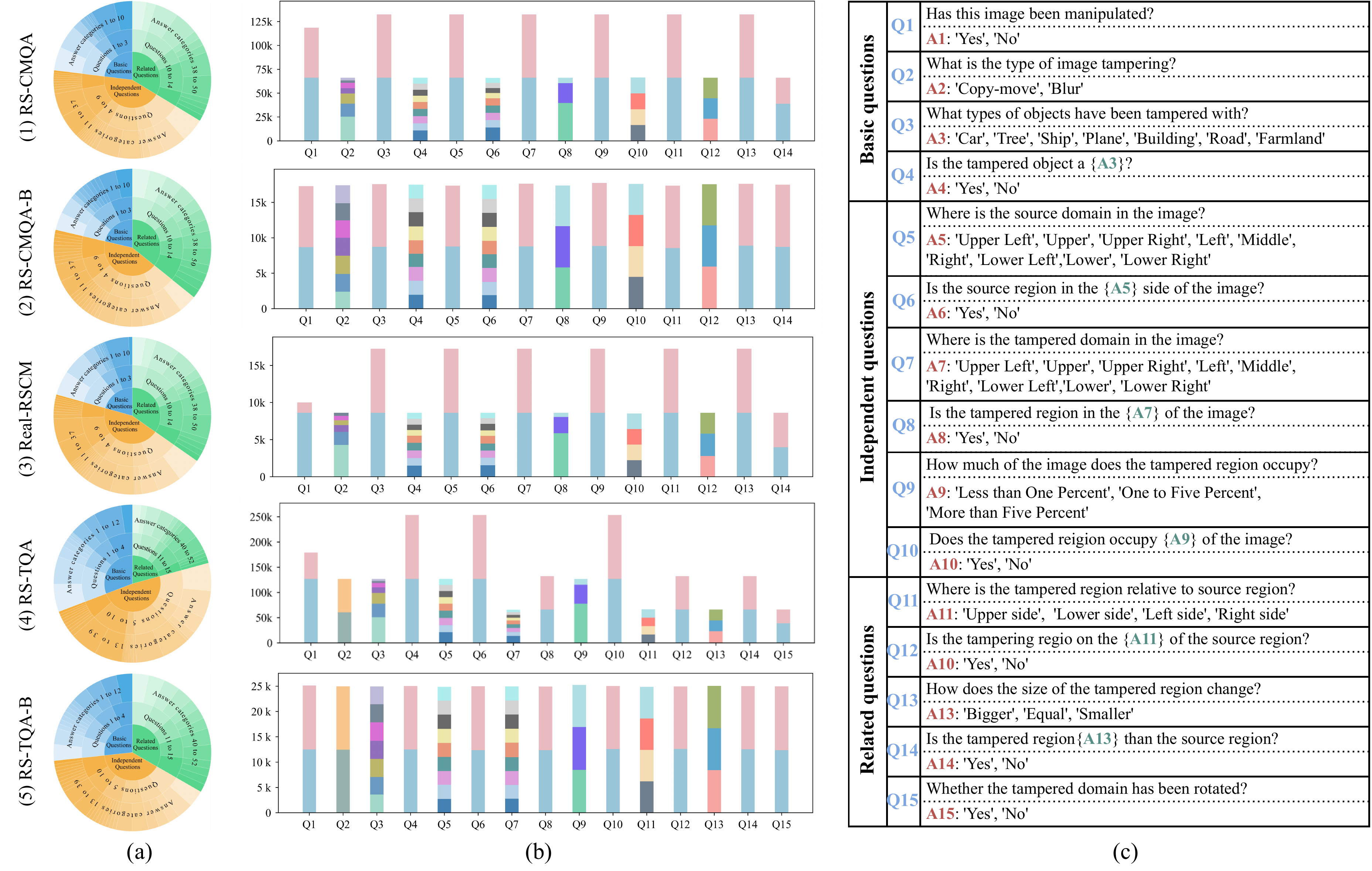}
    \caption{
(a) Distribution of basic, independent, and relational questions across the five datasets. (b) Detailed distribution of questions and answers in the five datasets. (c) Examples of question and answer types in the datasets.}
    \label{fig:qa}
\end{figure*}

The generation algorithm for CM-Q-A triplets is outlined in Algorithm \ref{alg:algorithm1}. Initially, raw images undergo manual preprocessing to ensure data quality. Tampered objects are randomly selected and scaled between $0.5\times$ and $1.5\times$, after which the modified object is placed at a random location within the image. To minimize excessive overlap between the source and tampered regions, the maximum overlap ratio is constrained to 5\% of the source region, ensuring that the source and tampered areas remain distinct. For the RS-TQA dataset, an additional blurred tampering algorithm is introduced. Selected objects are processed using one of three common blurring techniques: Gaussian blur, mosaic blur, or oil-painting smudge. For these blurred tampering cases, the source and tampered regions are considered identical. Through these algorithms and constraints, accurate tampered objects and their corresponding source and tampered regions are obtained.

To further reduce low-level artifacts introduced by direct copy-paste operations while preserving the intrinsic characteristics of copy-move forgery, two different refinement strategies are adopted for synthetic and realistic datasets. For the large-scale synthetic datasets (RS-CMQA, RS-TQA, RS-CMQA-B, and RS-TQA-B), a boundary-aware harmonization strategy is applied after the initial copy-move generation. Specifically, a transition band with a fixed width of 3 pixels is constructed around the tampered boundary using morphological dilation and erosion operations. Gaussian-weighted alpha blending is then performed within this transition band:
\[
I_{\mathrm{out}}=\alpha I_{\mathrm{tampered}}+(1-\alpha)I_{\mathrm{background}},
\]
where $\alpha\in[0,1]$ denotes the blending coefficient, which gradually decreases from the inner boundary to the outer boundary according to a Gaussian weighting function. Since the refinement is restricted to a narrow 3-pixel transition band, the copied object, tampering mask, and source-target correspondence remain unchanged while abrupt boundary transitions and local appearance inconsistencies are effectively reduced.

For the Real-RSCM dataset, we further introduce a localized diffusion-based refinement strategy using the FLUX.1 Fill model\cite{flux2024tools}. Unlike conventional diffusion-based image editing, the purpose of FLUX.1 Fill is not to regenerate manipulated content but to harmonize the transition area between the copied region and the surrounding background. Specifically, the original tampered region mask is first expanded by 20\% through mask dilation, and only the expanded region is provided to the diffusion model for refinement. To minimize unintended semantic modifications, we adopt highly conservative prompts. A conservative prompt is adopted: "Refine the selected region to make the transition with surrounding areas more natural. Preserve the original objects, structures, and spatial layout as much as possible. Avoid adding, removing, or changing semantic content. The negative prompt is new objects, object removal, structural changes, scene modification, content generation, geometric distortion."

Multiple refinement results are generated for each candidate image. All generated candidates are manually inspected according to three criteria: (1) preservation of the copied object identity, (2) preservation of the intended manipulation location, and (3) absence of newly introduced artifacts. Only the visually most realistic results are retained. This refinement process effectively suppresses low-level artifacts while maintaining the spatial characteristics, source-target correspondence, and semantic consistency required by copy-move forgery.

Questions and answers are automatically generated based on each step of the tampering process. For each tampered instance, the dataset provides the tampered image, original image, segmentation mask, source region mask, and tampered region mask. An example of dataset images is presented in Figure \ref{fig:example}, while specific question-answer pairs are illustrated in Figure \ref{fig:qa}(c).

For all datasets, 70\% of the data is allocated to the training set, while 15\% is assigned to both the validation and test sets. RS-CMQA, RS-CMQA-B, and Real-RSCM contain 14 question categories and 51 answer types. In contrast, RS-TQA and RS-TQA-B additionally incorporate tampering type classification. Specifically, the question "What is the type of image tampering?" is exclusive to these two datasets. All questions are categorized into basic, independent, and related questions, with their distribution across the five datasets illustrated in Figure \ref{fig:qa}(a). The detailed distribution of questions and answers is presented in Figure \ref{fig:qa}(b).

To facilitate standardized evaluation, the proposed benchmarks formulate question answering over a predefined answer vocabulary. This design is intentionally adopted to provide an objective, reproducible, and quantitatively comparable evaluation protocol, following the common practice in existing remote sensing visual question answering (RSVQA) benchmarks. Compared with free-form textual generation, the closed-set formulation avoids evaluation ambiguity caused by diverse yet semantically equivalent responses while enabling fair comparison across different methods.

Although the final answers are selected from a predefined vocabulary, obtaining the correct answer still requires integrating visual forgery evidence with textual queries. The predefined answers are designed to cover the major attributes involved in copy-move forgery analysis, including tampering existence, object category, spatial location, relative position, and object size. Accordingly, the proposed benchmark is intended to evaluate structured forgery-oriented multimodal reasoning under a unified evaluation protocol, rather than open-ended natural language generation.

\vspace{-5pt}
\subsection{RS-CMQA and RS-CMQA-B Datasets}
The RS-CMQA dataset comprises 118k images and 1.3 million CM-Q-A triplets, serving as a foundational training resource and evaluation benchmark for remote sensing cross-modal question answering (RSCMQA). The distributions of questions and answers are illustrated in Figure \ref{fig:qa}(a)(1) and Figure \ref{fig:qa}(b)(1). While RS-CMQA addresses the absence of prior large-scale datasets and enables models to acquire extensive domain knowledge, it exhibits significant imbalance across question types, which may introduce bias into question-answering tasks. This imbalance presents both opportunities and challenges for advancing RSCMQA research.

To alleviate the long-tail distribution issue and provide researchers with a more equitable evaluation resource, we construct RS-CMQA-B, a balanced subset of RS-CMQA generated through weighted random sampling of all questions (where B denotes “balance”). RS-CMQA-B contains 245k CM-Q-A triplets, with an average of 17.5k triplets per question type. The variation in question counts across categories does not exceed 2\%, and the distribution of answers within each question type remains balanced. As shown in Figure \ref{fig:qa}(a)(2) and Figure \ref{fig:qa}(b)(2), RS-CMQA-B constitutes a substantial and high-quality dataset, offering a fair and balanced evaluation benchmark for RSCMQA research.

\vspace{-3pt}
\subsection{RS-TQA and RS-TQA-B Datasets}
The RS-TQA dataset extends RS-CMQA by incorporating blurred tampering, comprising 179k images and 2.1 million T-Q-A triplets, where T denotes Tampering. It includes two types of tampering: copy-move and blurred tampering. The distributions of questions and answers are illustrated in Figure \ref{fig:qa}(a)(4) and Figure \ref{fig:qa}(b)(5). RS-TQA enables the evaluation of model robustness and transferability when confronted with alternative tampering techniques, thereby providing a more comprehensive assessment of models designed for the RSCMQA task. However, similar to RS-CMQA, RS-TQA exhibits imbalance across question types, which may bias evaluation outcomes.

To address this limitation, we construct RS-TQA-B, a balanced subset of RS-TQA generated through weighted random sampling of all questions. RS-TQA-B contains 375k CM-Q-A triplets, with an average of 25k triplets per question type. The variation in question counts across categories does not exceed 2\%, and the distribution of answers within each question type remains similarly balanced. As shown in Figure \ref{fig:qa}(a)(5) and Figure \ref{fig:qa}(b)(5), RS-TQA-B constitutes a substantial and high-quality dataset, offering an expanded yet fair evaluation benchmark for RSCMQA research.

\begin{figure}[!t]
    \centering
    \includegraphics[width=\linewidth]{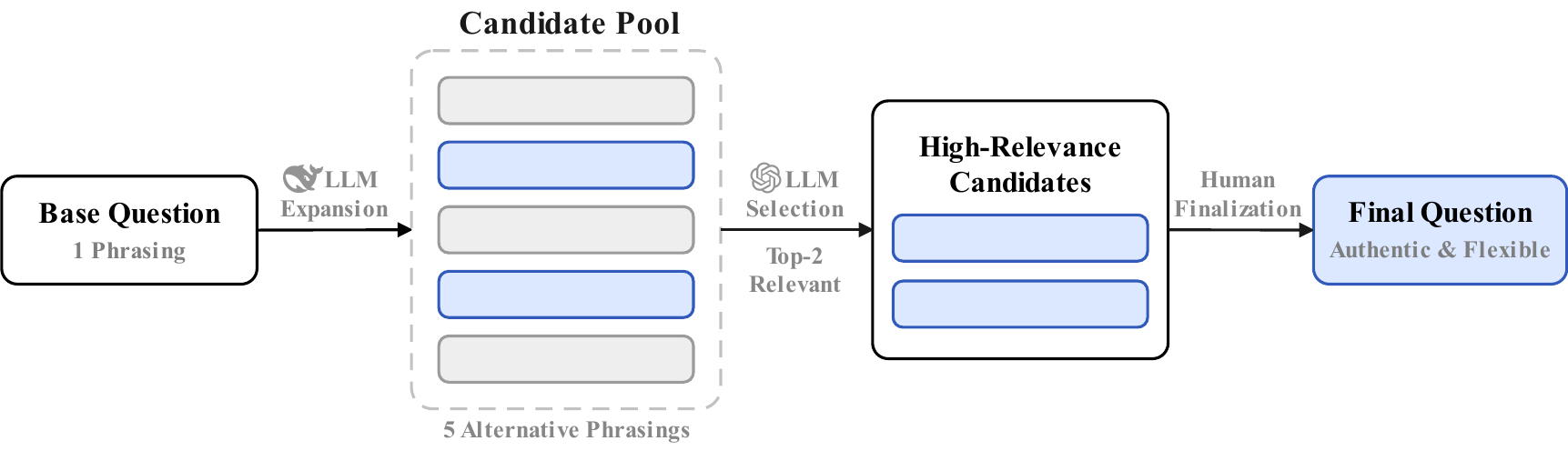}
    \caption{Pipeline of question expansion, LLM filtering, and human refinement for Real-RSCM annotations.}
    \label{fig:questions}
\end{figure}

\begin{figure}[!t]
    \centering
    \includegraphics[width=\linewidth]{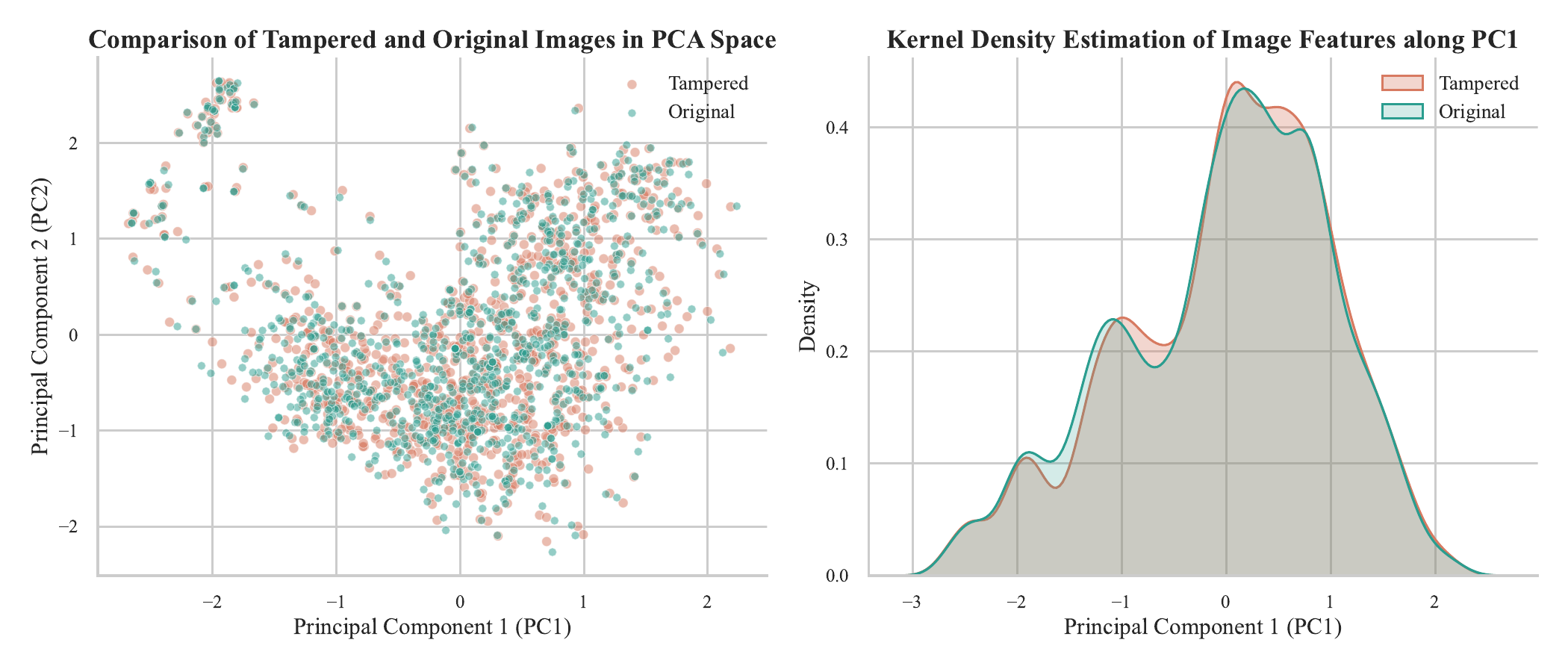}
    \caption{Comparison of tampered and original images in PCA space, with kernel density estimation along PC1 highlighting distributional differences.}
    \label{fig:pca}
\end{figure}

\vspace{-5pt}
\subsection{Real-RSCM dataset}

To overcome the limitations of rule-based dataset generation, which often produces easily detectable tampering, we constructed Real-RSCM, a realistic and challenging dataset tailored for remote sensing copy-move forgery perception. The distribution of questions and answers is illustrated in Figure \ref{fig:qa}(a)(3) and Figure \ref{fig:qa}(b)(3). Real-RSCM contains 10,000 images and 173,000 CM-Q-A triplets, designed through a multi-stage process to ensure authenticity, diversity, and annotation accuracy.

\begin{figure}[!t]
    \centering
    \includegraphics[width=0.9\linewidth]{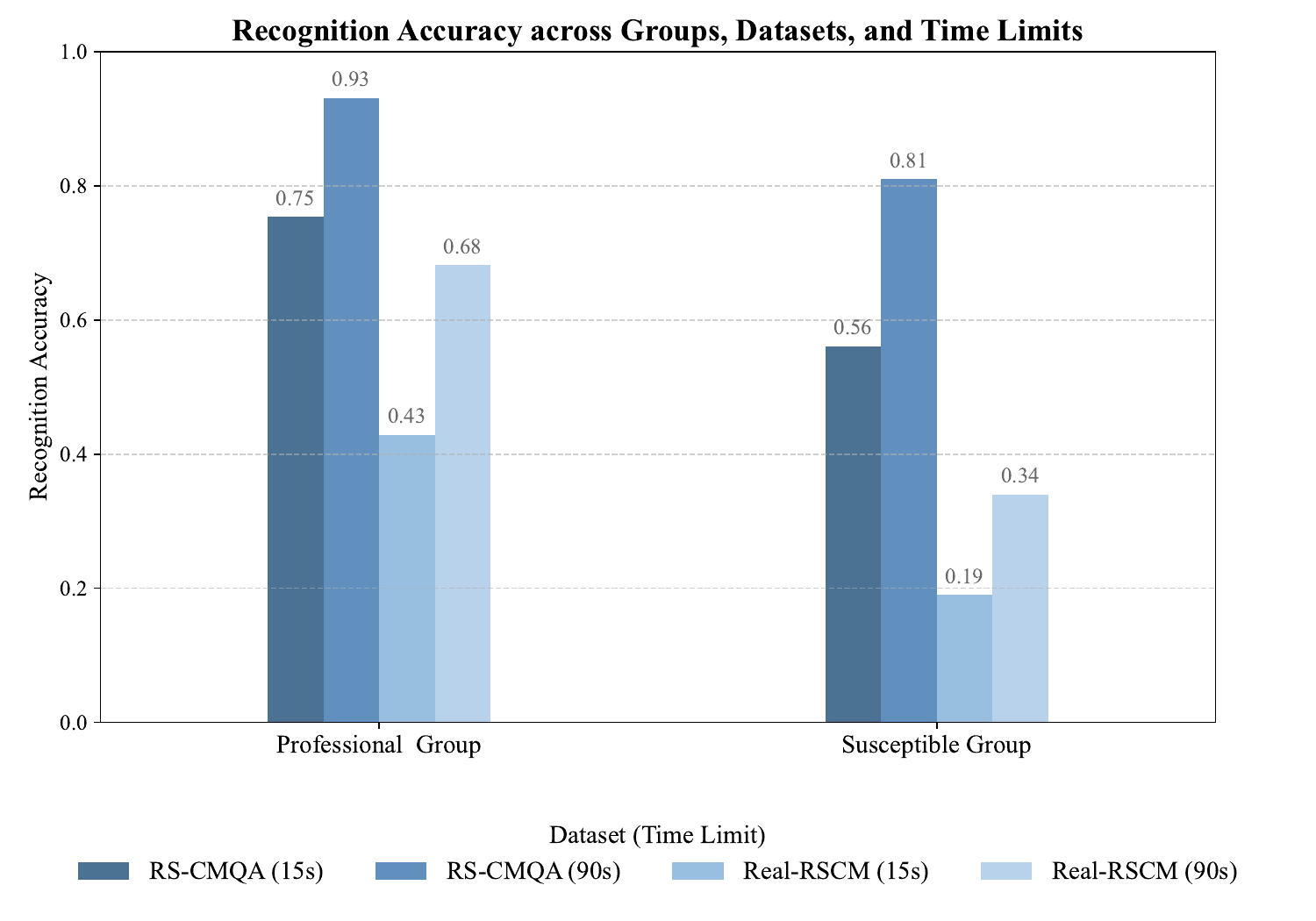}
    \caption{Comparison of recognition accuracy across participant groups, datasets, and time limits.}
    \label{fig:people}
\end{figure}

All tampering instances are manually annotated with strict constraints on region size and spatial placement, thereby avoiding unrealistic overlaps or distortions. Each tampered object undergoes automated quality checks and human verification, ensuring semantic clarity, spatial plausibility, and question–answer consistency. To further enhance dataset difficulty, we performed two rounds of reconstruction: in each round, the 1,000 images with the highest recognition accuracy were removed and replaced with 1,000 newly annotated samples, ensuring that the dataset remains challenging and diverse.

For question design, we adopted an LLM-assisted generalization and human selection process, which is shown in Figure \ref{fig:questions}. Each question was expanded into five alternative phrasings, from which large language models selected the two most relevant. Human annotators then finalized one phrasing, significantly improving the authenticity, flexibility, and accuracy of the annotations.

\subsection{Datasets Evaluation}
To systematically evaluate recognition accuracy under diverse demographic and temporal constraints, we conducted controlled experiments on the RS-CMQA and Real-RSCM datasets. Participants were stratified into two cohorts: (1) a professional group comprising 20 researchers with advanced educational backgrounds and specialized expertise in remote sensing image analysis, including segmentation and tampering detection; and (2) a susceptible group consisting of 20 randomly selected individuals aged 45 years or older without higher education. Informed by prior studies reporting that the average viewing duration of short-form videos is approximately 10–15 seconds\cite{gong2024relationship,jiang2025losing}, we established 15 seconds as the rapid browsing condition and 90 seconds as the careful inspection condition. Each participant was tasked with examining 20 randomly sampled tampered images from both datasets under both temporal constraints and delineating the manipulated regions.

As depicted in Figure \ref{fig:people}, although RS-CMQA is comparatively easier to discern, rapid browsing nonetheless induced misjudgments, particularly among the susceptible group. This observation underscores the potential role of vulnerable populations in inadvertently facilitating the dissemination of misinformation. While professionals attained relatively high accuracy given extended inspection time, the associated cognitive and temporal demands render such exhaustive evaluation impractical in real-world applications.

By contrast, the Real-RSCM dataset presented a substantially greater challenge. Even professional participants demonstrated a pronounced decline in recognition accuracy, whereas the susceptible group was largely unable to produce reliable judgments. Under rapid browsing, their accuracy was limited to 19\%, and even with extended inspection, performance improved only to 34\%. These findings highlight the inherent difficulty of detecting subtle manipulations and emphasize the elevated risk of misinformation propagation among non-expert audiences.

Principal Component Analysis (PCA) provides a compact representation of high-dimensional image data by projecting it into a lower-dimensional space while preserving the most significant variance. To evaluate the Real-RSCM dataset, we compared the distributions of original and tampered images in PCA space and further examined kernel density estimates along the first principal component. As shown in Figure \ref{fig:pca}, the differences between the two distributions are minimal, with nearly identical density profiles. This observation indicates that the tampered images in Real-RSCM maintain a similar global image distribution to the original remote sensing images, suggesting that the dataset preserves the overall visual characteristics of real remote sensing scenes.

\section{Methodology}
\label{sec:method}

\begin{figure*}[!t]
    \centering
    \includegraphics[width=\textwidth]{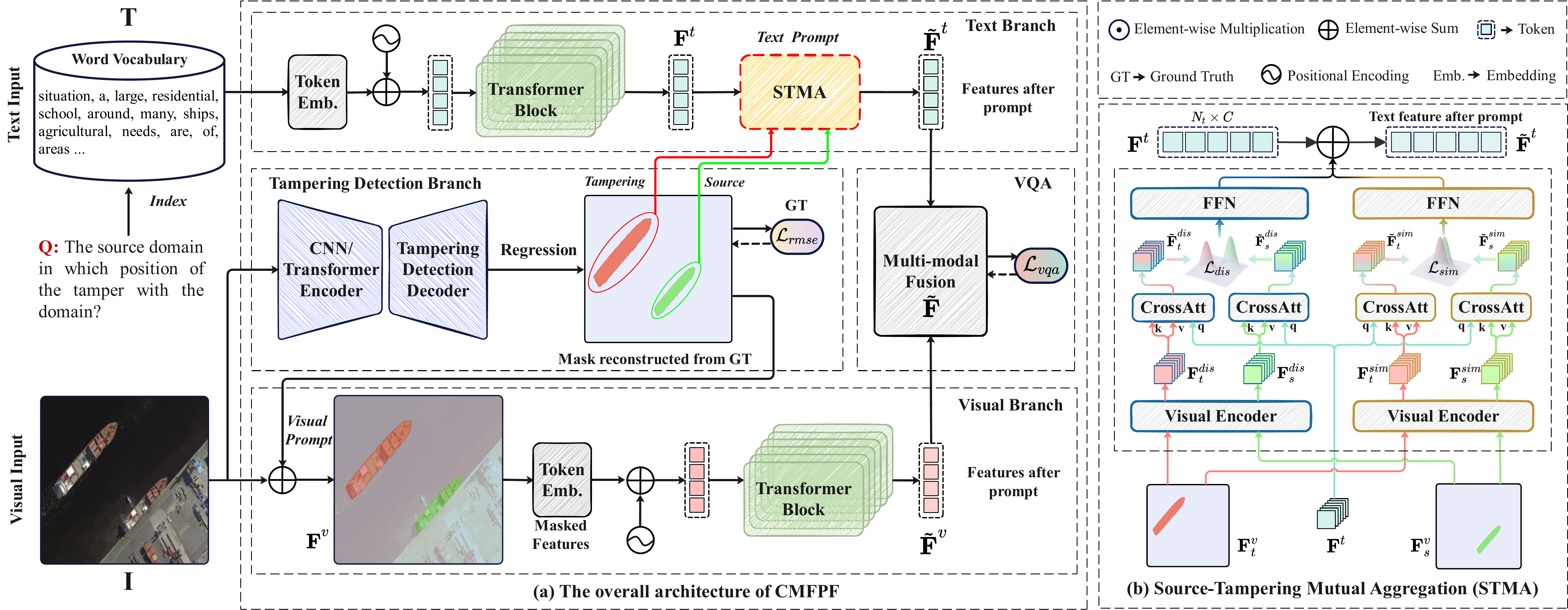}
    \caption{
    An illustration of the proposed CMFPF framework and the STMA module, which provides tampering prompts for the textual modality.
    }
    \label{fig:framework}
\end{figure*}

\label{sec:4}
To identify the source and tampered regions and facilitate relevant reasoning, we propose the Region-Discrimination-Guided Multimodal Copy-Move Forgery Perception Framework (CMFPF). The CMFPF involves a two-phase training process: (1) Training the tampering detection network to generate visual and textual prompts; and (2) Leveraging the multimodal representations of these prompts for reasoning and response. For the tampering detection network, masks of the source and tampered regions serve as ground truth to train the visual branch, while the network outputs are utilized as prompts for the VQA network. The overall architecture of the CMFPF is shown in Figure \ref{fig:framework}(a).

\subsection{Tampering Detection for Visual Prompt}
In scenarios containing potential tampered regions, we utilize a pixel-level reconstruction network to provide fine-grained guidance for downstream question-answering tasks. Given an input image $\mathbf{I} \in \mathbb{R}^{H \times W \times 3}$, the tampering detection decoder (TDD) outputs the source region reconstruction mask \( \mathbf{F}^v_s \) and the tampered region reconstruction mask \( \mathbf{F}^v_t \):
\begin{equation}
[\mathbf{F}^{v}_{s}, \mathbf{F}^{v}_{t}]=\operatorname{TDD}(\mathbf{I}).
\end{equation}

Since both the original image and the masks belong to the same visual modality, for prompts in the visual modality, we directly average \( \mathbf{F}^v_s \) and \( \mathbf{F}^v_t \), then overlay them onto the original image:
\begin{equation}
\mathbf{F}^v = \mathbf{I} \oplus \operatorname{Avg}\left(\mathbf{F}^v_s \oplus \mathbf{F}^v_t \right).
\end{equation}

Here, \( \oplus \) represents element-wise addition, \( \mathbf{F}^v \) represents the visually prompted image, which is processed through the visual encoder to obtain the final visual feature \( \tilde{\mathbf{F}}^v \).

\begin{equation}
\tilde{\mathbf{F}}^v = \operatorname{VisualEncoder}\left( \mathbf{F}^v \right).
\end{equation}

As shown in Figure \ref{fig:framework}(a), both the visual encoder and text encoder are referred to as Transformer-based encoding modules.

\subsection{Tampering Detection for Text Prompts}
For the textual modality, the input question $\mathbf{Q}$ is first processed through word indexing and token embedding, followed by the text encoder to generate the textual feature \( \mathbf{F}^t \):
\begin{equation}
\mathbf{F}^t = \operatorname{TextEncoder}\left( \operatorname{Emb}\left(\operatorname{Ind}\left( \mathbf{Q} \right)\right)\right).
\end{equation}

Under the guidance of source and tampered masks, the \textit{Source-Tampering Mutual Aggregation (STMA)} module serves as the key bridge between tampering detection and question answering. Unlike conventional approaches that only localize tampered regions, STMA processes the source and tampered masks through two independent encoders to generate complementary feature representations. One branch emphasizes differential information, capturing fine-grained discrepancies between source and tampered regions, while the other highlights associative information, modeling their correlations and similarities. By explicitly modeling both aspects, STMA injects forgery-aware cues into the textual modality via Cross-Attention, ensuring that the language branch is contextually guided by pixel-level tampering evidence. This design effectively transforms detection outputs into reasoning signals, enabling fine-grained semantic and spatial analysis in copy-move question answering. The module’s structure is illustrated in Figure \ref{fig:framework}(b).
Specifically, the tampered region reconstruction mask \( \mathbf{F}^v_t \) and the source region reconstruction mask \( \mathbf{F}^v_s \), generated by the forgery detection network, are processed through two distinct image encoders with non-shared parameters, producing \( \mathbf{F}^{dis}_t \), \( \mathbf{F}^{dis}_s\), \( \mathbf{F}^{sim}_t \), and \( \mathbf{F}^{sim}_s \).
\begin{subequations}
   \begin{align}
      \mathbf{F}^{dis }_t, \mathbf{F}^{dis}_s &= \operatorname{VisualEncoder_1}(\mathbf{F}^v_t, \mathbf{F}^v_s),\\
      \mathbf{F}^{sim}_t, \mathbf{F}^{sim}_s &= \operatorname{VisualEncoder_2}(\mathbf{F}^v_t, \mathbf{F}^v_s), 
   \end{align}
\end{subequations}
where \( \mathbf{F}^{dis }_s \) and \( \mathbf{F}^{dis}_t \) are subsequently utilized to extract discriminative information between the source and target regions, while \( \mathbf{F}^{sim}_s \) and \( \mathbf{F}^{sim}_t \) are employed to capture the relational information between these regions.

The textual feature \( \mathbf{F}^t \) undergoes Cross-Attention operations with \( \mathbf{F}^{dis}_t \), \( \mathbf{F}^{dis}_s \), \( \mathbf{F}^{sim}_t \), and \( \mathbf{F}^{sim}_s \), where the embedding of the textual feature serves as the query, while the embeddings of the image features are used as the key and value. This process yields the forgery-related features \( \tilde{\mathbf{F}}_{t}^{dis} \), \( \tilde{\mathbf{F}}_{s}^{dis} \), \( \tilde{\mathbf{F}}_{t}^{sim} \), and \( \tilde{\mathbf{F}}_{s}^{sim} \), combined with textual information. C is the feature vector dimension in the following formula.

\begin{subequations}
    \begin{align}
   \tilde{\mathbf{F}}_{t}^{dis} &= \operatorname{Softmax} \left( \frac{\operatorname{Emb}\left( \mathbf{F}^t \right)\operatorname{Emb}\left(( \mathbf{F}_{t}^{dis})^T\right)}{\sqrt{C}} \right)\operatorname{Emb}( \mathbf{F}_{t}^{dis}) \\
   \tilde{\mathbf{F}}_{s}^{dis} &= \operatorname{Softmax} \left( \frac{\operatorname{Emb}\left( \mathbf{F}^t \right)\operatorname{Emb}\left(( \mathbf{F}_{s}^{dis})^T\right)}{\sqrt{C}} \right)\operatorname{Emb}( \mathbf{F}_{s}^{dis}) \\
   \tilde{\mathbf{F}}_{t}^{sim} &= \operatorname{Softmax} \left( \frac{\operatorname{Emb}\left( \mathbf{F}^t \right)\operatorname{Emb}\left(( \mathbf{F}_{t}^{sim})^T\right)}{\sqrt{C}} \right)\operatorname{Emb}( \mathbf{F}_{t}^{sim}) \\
   \tilde{\mathbf{F}}_{s}^{sim} &= \operatorname{Softmax} \left( \frac{\operatorname{Emb}\left( \mathbf{F}^t \right)\operatorname{Emb}\left(( \mathbf{F}_{s}^{sim} )^T\right)}{\sqrt{C}} \right)\operatorname{Emb}( \mathbf{F}_{s}^{sim})
    \end{align}
\end{subequations}

The discriminative information within \( \tilde{\mathbf{F}}_{t}^{dis} \) and \( \tilde{\mathbf{F}}_{s}^{dis} \) is extracted as the textual modality's tampered region difference feature embeddings, while the similarity information within \( \tilde{\mathbf{F}}_{t}^{sim} \) and \( \tilde{\mathbf{F}}_{s}^{sim} \) is extracted as the textual modality's tampered region similarity feature embeddings. These features are integrated into the original textual features \( \mathbf{F}^t \) through a three-layer feedforward neural network, resulting in the prompted textual feature \( \tilde{\mathbf{F}}^{t} \):
\begin{equation}
\tilde{\mathbf{F}}^t = \mathbf{F}^t \oplus \operatorname{FFN}(\operatorname{Dis}(\tilde{\mathbf{F}}_{t}^{dis}, \tilde{\mathbf{F}}_{s}^{dis})) \oplus \operatorname{FFN}(\operatorname{Sim}(\tilde{\mathbf{F}}_{t}^{sim}, \tilde{\mathbf{F}}_{s}^{sim}))
\end{equation}
where the differences and similarities of the features are both evaluated using the Kullback–Leibler (KL) divergence.

\begin{figure}[!t]
    \centering
    \includegraphics[width=\linewidth]{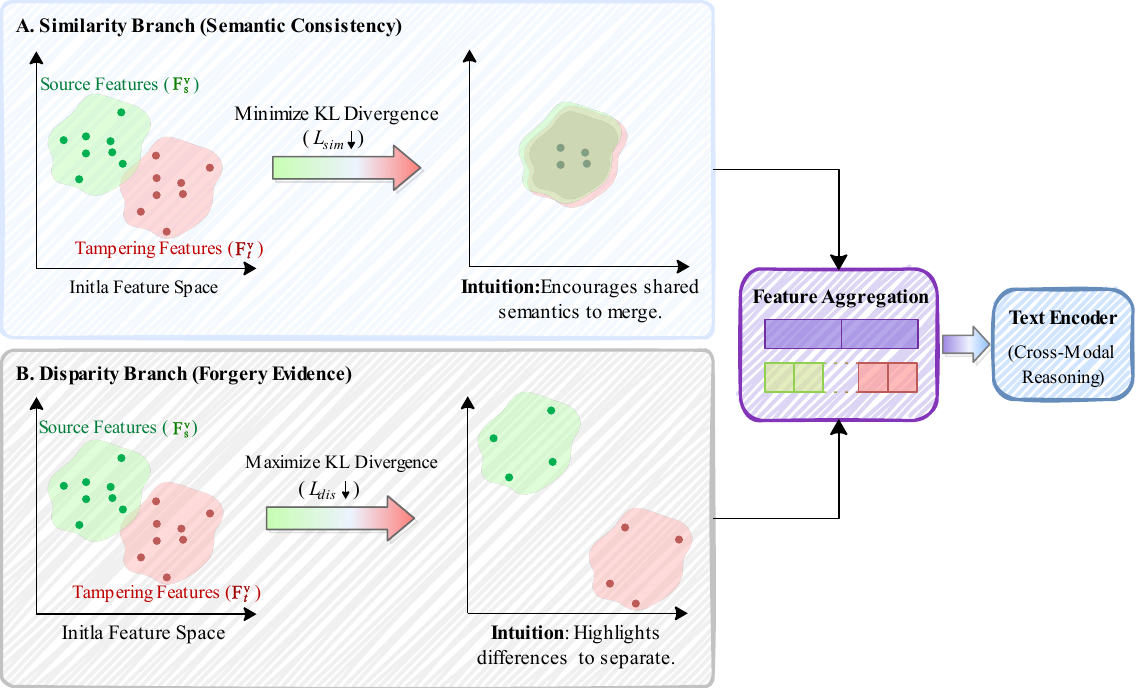}
    \caption{Intuitive illustration of the Source–Tampering Mutual Aggregation (STMA) mechanism.
The module adopts a dual‑branch design that imposes strong contrastive constraints on the feature space. A. Similarity branch: minimizes the KL divergence to pull the source and tampered feature distributions closer, capturing shared semantic content. B. Dissimilarity branch: maximizes the KL divergence to push the distributions apart, highlighting region-level discrepancies indicative of tampering. The complementary signals from both branches are aggregated to generate discriminative prompts for the text encoder.}
    \label{fig:stma1}
\end{figure}

Intuitively, the STMA module imposes a dual contrastive constraint on the feature distributions of the source and tampered regions. The \emph{similarity branch} focuses on cases where the two regions share consistent semantics (e.g., both depict ``ships'') and encourages their feature distributions to move closer in the latent space, thereby preserving the core object identity across domains. In contrast, the \emph{disparity branch} emphasizes subtle discrepancies---such as interpolation artifacts or contextual inconsistencies---and drives the corresponding distributions apart. As illustrated in Figure~\ref{fig:stma1}, this pull-push mechanism reshapes the feature space so that shared semantics are compactly clustered while tampering cues are explicitly separated. The aggregated signals are then injected as discriminative cross-modal prompts into the text encoder, enabling more precise reasoning about both the object content and the manipulation traces.

Although STMA adopts a dual-branch pull–push constraint, it is fundamentally different from conventional dual-branch contrastive learning frameworks. Existing contrastive learning methods mainly optimize the similarity or discrepancy between independent samples to obtain general discriminative representations. In contrast, STMA focuses on the intrinsic relationship between the source region and the tampered region within the same manipulated image. The similarity branch preserves semantic consistency between the copied object and its source, while the disparity branch highlights manipulation-induced differences, such as boundary artifacts, contextual inconsistency, and spatial transformation cues.

Moreover, STMA is not designed as a generic representation alignment module. As shown in Figure 8(b), the source and tampered masks are processed by two non-shared encoders and then interacted with textual features through four Cross-Attention operations. The resulting similarity-aware and discrepancy-aware features are further transformed into cross-modal prompts and injected into the text branch. Therefore, STMA serves as a detection-to-reasoning bridge that converts pixel-level forgery evidence into textual reasoning cues for VQA, rather than merely producing enhanced embeddings for recognition or retrieval.

Finally, the prompted visual and textual representations are fused to perform the questio  n-answering task: 
\begin{equation}
\tilde{\mathbf{F}} = \operatorname{FFN}(\operatorname{Mul}(\tilde{\mathbf{F}}^t \text{,}  \tilde{\mathbf{F}}^v)),
\end{equation}
where \( \text{Mul} \) denotes element-wise multiplication, and \( \text{FFN} \) refers to a feedforward neural network comprising three fully connected layers and three activation layers. The resulting multimodal feature \( \tilde{\mathbf{F}}\) is used for VQA prediction.

\subsection{Loss Function}
The loss function  $\mathcal{L}$ consists of the tampering detection loss, the VQA loss, and the feature metric loss. The reconstruction loss for forgery detection is derived based on the Root Mean Square Error (RMSE), while the VQA loss is determined using Cross-Entropy (CE) loss. The feature metric loss is calculated through the Kullback–Leibler (KL) divergence. RMSE quantifies the differences between the predicted source region mask and tampered region mask against the ground truth. Specifically, RMSE loss is given by:
\begin{equation}
\mathcal{L}_{\mathit{rmse}} = \sqrt{\frac{1}{n} \sum_{i=1}^{n} (\hat{\mathbf{F}}^v_t - \mathbf{F}^v_t)^2}+ \sqrt{\frac{1}{n} \sum_{i=1}^{n} (\hat{\mathbf{F}}^v_s - \mathbf{F}^v_s)^2},
\end{equation}
where \( n \) represents the number of samples, \( \hat{\mathbf{F}}^v_t \) and \( \hat{\mathbf{F}}^v_s \) represent the ground truth masks for the tampered region and the source region, respectively, while \( \mathbf{F}^v_t \) and \( \mathbf{F}^v_s \) correspond to the tampered region and source region masks output by the forgery detection network.

The Cross-Entropy Loss for VQA is expressed as:
\begin{equation}
\mathcal{L}_{\mathit{vqa}} = -\frac{1}{n} \sum_{i=1}^{n} y_i \log(\hat{y}_i),
\end{equation}
where \( y_i \) denotes the ground truth answer and $\hat{y}_i$ represents the probability predicted through the fused representation $\tilde{\mathbf{F}}$.

The formula for KL divergence is as follows, where P(i) and Q(i) are the feature distributions after softmax normalization:
\begin{equation}
\mathit{D}_{kl}(P \| Q) = \sum_{i} P(i) \cdot \log\left(\frac{P(i)}{Q(i)}\right).
\end{equation}
The feature metric loss \( L_{KL} \) comprises two components: the similarity loss \( L_{sim} \) and the discriminative loss \( L_{dis} \). These are defined respectively by the KL divergence and the reciprocal of the KL divergence:
\begin{subequations}
   \label{eq:eq12}
   \begin{align}
\mathcal{L}_{\mathit{sim}} &= \mathit{D}_{kl}(\tilde{\mathbf{F}}_{t}^{sim} \| \tilde{\mathbf{F}}_{s}^{sim}), \\
\mathcal{L}_{\mathit{dis}} &= \frac{1}{\mathit{D}_{kl}(\tilde{\mathbf{F}}_{t}^{dis} \| \tilde{\mathbf{F}}_{s}^{dis})+\sigma}, \\
\mathcal{L}_{\mathit{kl}} &= \mathcal{L}_{\mathit{sim}} + \mathcal{L}_{\mathit{dis}},
   \end{align}
\end{subequations}
where \( \sigma \) represents a tiny positive constant introduced to prevent division by zero anomalies.

Although the KL-divergence-based formulation of STMA may appear mathematically involved, its underlying principle is a simple pull-push mechanism in the latent feature space. As shown in Figure~\ref{fig:stma1}, the \emph{similarity loss} $\mathcal{L}_{\text{sim}}$ minimizes the KL divergence between the source and tampered feature distributions when they are semantically consistent, effectively pulling them closer and stabilizing the shared object identity. Conversely, the \emph{disparity loss} $\mathcal{L}_{\text{dis}}$, defined as the reciprocal of the KL divergence, encourages larger divergence when subtle but meaningful differences exist, thereby amplifying forgery cues such as boundary artifacts or mismatches between the pasted copy and its original context. This joint optimization yields highly discriminative prompts: similarity informs the text branch about \emph{what} the object is, while disparity highlights \emph{where} and \emph{how} it has been manipulated. The ablation study (Table \ref{tab:stma_ablation}) empirically confirms that removing either the pull or push constraint leads to a notable performance drop, underscoring the necessity of the dual-branch design.

The overall loss $\mathcal{L}$ is defined as follows:
\begin{equation}
\mathcal{L} = \alpha \cdot \mathcal{L}_{\mathit{vqa}} + (1 - \alpha) \cdot \mathcal{L}_{\mathit{rmse}} + \mathcal{L}_{\mathit{kl}}.
\end{equation}
where \(\alpha\) is a trade-off coefficient, balancing the weights of the forgery detection loss and the VQA loss. The feature metric loss stabilizes rapidly to a negligible value after training begins; therefore, no specific adjustments are applied to \(\mathcal{L}_{\mathit{kl}}\).

\section{Experiments}

\begin{figure}[!t]
    \centering
    \includegraphics[width=1\linewidth]{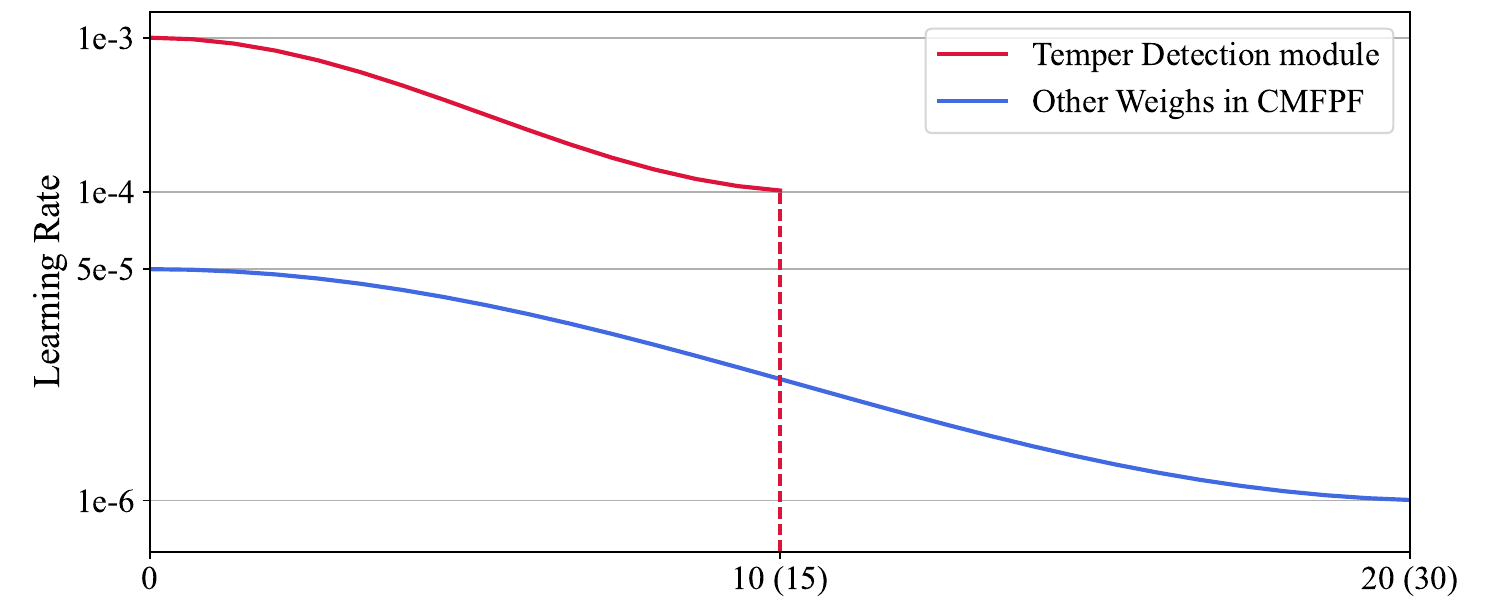}
    \caption{Learning Rate Configuration in CMFPF Training. The model undergoes 20 epochs of training on the RS-CMQA and RS-TQA datasets, while it is trained for 30 epochs on the RS-CMQA-B, Real-RSCM, and RS-TQA-B datasets. The tamper detection module updates its parameters exclusively during the initial half of the training process.}

    \label{fig:lr}
\end{figure}

\noindent \textbf{Evaluation metrics.}
The overall accuracy (OA) across all questions serves as an intuitive measure to evaluate the model's prediction performance. The average accuracy (AA) across different question categories assesses the model's performance balance, while the accuracy of individual question types provides a more detailed evaluation. All metrics are expressed as percentages.
\noindent \textbf{Experimental settings.} 
All models were trained for 20 epochs on the RS-CMQA and RS-TQA datasets, and for 30 epochs on the RS-CMQA-B, Real-RSCM, and RS-TQA-B datasets. The batch size was set to 32, and the Adam optimizer was employed. The hyperparameter $\alpha$ was fixed at 0.7. \textbf{Text and visual heads.} The text head and visual head of CMFPF are based on the CLIP-pretrained model~\cite{radford2021learning}. The text encoder is a 12-layer Transformer with hidden size 512, intermediate size 2048, and 8 attention heads, using GELU activation. The vision encoder is a 12-layer ViT-B with hidden size 768, intermediate size 3072, 12 attention heads, and patch size 32, operating on $224\times224$ input images. Both encoders project into a 512-dimensional multimodal space. \textbf{Tampering detection module.} This component was implemented using a U-Net architecture with four down-sampling and four up-sampling stages. The encoder path consists of DoubleConv blocks with channel sizes 64–128–256–512–1024, followed by symmetric up-sampling layers with skip connections. The final output layer produces tampering masks. \textbf{Learning rate schedule.} As illustrated in Figure~\ref{fig:lr}, we adopted a cosine annealing decay strategy. For the tampering detection module, the learning rate decreased from $1\times10^{-3}$ to $1\times10^{-4}$, with parameter updates restricted to the first half of training. For the remaining parameters, the learning rate decayed from $5\times10^{-4}$ to $1\times10^{-6}$. To ensure fairness, all baseline models leveraged pretrained encoders. The learning rate for CNN-based baselines decreased from $1\times10^{-3}$ to $1\times10^{-4}$, whereas for transformer-based baselines, it decayed from $5\times10^{-4}$ to $1\times10^{-6}$. \textbf{Implementation details.} All experiments were conducted on four NVIDIA RTX 4090 GPUs, utilizing PyTorch version~2.3.0 and CUDA version~12.1.

\begin{table*}[t]
\centering
\caption{
Evaluation with state-of-the-art methods on the RS-CMQA test set, RS-CMQA-B test set and Real-RSCM test set, with best metrics highlighted in \textbf{bold}.
}
\vspace{-6pt}
\resizebox{1\textwidth}{!}{
\renewcommand{\arraystretch}{1.1}
\large 
\begin{tabular}{c|c|ccc|cccccc|ccccc|cc}
\midrule[0.9pt]
 & & \multicolumn{3}{c|}{Basic Questions} & \multicolumn{6}{c|}{Independent Questions} & \multicolumn{5}{c|}{Related Questions} &  &  \\
 & \multirow{-2}{*}{Method} & Q1 & Q2 & Q3 & Q4 & Q5 & Q6 & Q7 & Q8 & Q9 & Q10 & Q11 & Q12 & Q13 & Q14 & \multirow{-2}{*}{OA} & \multirow{-2}{*}{AA} \\ \midrule[0.9pt]
\multirow{14}{*}{\rotatebox{90}{RS-CMQA Test Set}} & \multicolumn{1}{l|}{*General VQA Methods} & \multicolumn{1}{l}{} & \multicolumn{1}{l}{} & \multicolumn{1}{l|}{} & \multicolumn{1}{l}{} & \multicolumn{1}{l}{} & \multicolumn{1}{l}{} & \multicolumn{1}{l}{} & \multicolumn{1}{l}{} & \multicolumn{1}{l|}{} & \multicolumn{1}{l}{} & \multicolumn{1}{l}{} & \multicolumn{1}{l}{} & \multicolumn{1}{l}{} & \multicolumn{1}{l|}{} & \multicolumn{1}{l}{} & \multicolumn{1}{l}{} \\
& SAN (CVPR, 2016)\cite{yang2016stacked} & 91.26 & 98.36 & 99.28 & 49.77 & 81.80 & 90.23 & 96.57 & 97.86 & 97.51 & 82.72 & 91.22 & 77.06 & 88.51 & 64.49 & 88.22 & 86.19 \\
& MAC (ICLR, 2018)\cite{hudson2018compositional} & 87.73 & 98.10 & 99.65 & 58.29 & 84.21 & 89.05 & 96.64 & 96.83 & 98.57 & 87.97 & 93.97 & 83.63 & 90.69 & 64.97 & 89.64 & 87.88 \\
& MCAN (CVPR, 2019)\cite{yu2019deep} & 67.71 & 96.38 & 98.45 & 36.29 & 74.81 & 54.93 & 83.56 & 94.37 & 97.23 & 58.30 & 77.62 & 69.06 & 81.75 & 57.82 & 77.70 & 74.88 \\
& DVQA (NeurIPS, 2021)\cite{wen2021debiased} & 87.61 & 97.91 & 99.67 & 50.42 & 82.06 & 88.73 & 94.41 & 95.90 & 98.24 & 84.70 & 92.52 & 69.06 & 81.01 & 63.75 & 86.75 & 84.71 \\
& BLIP-2 (ICML, 2023)\cite{li2023blip}  & \multicolumn{1}{l}{87.23} & \multicolumn{1}{l}{97.18} & \multicolumn{1}{l|}{98.19} & \multicolumn{1}{l}{37.00} & \multicolumn{1}{l}{72.83} & \multicolumn{1}{l}{80.34} & \multicolumn{1}{l}{91.99} & \multicolumn{1}{l}{94.08} & \multicolumn{1}{l|}{96.23} & \multicolumn{1}{l}{66.98} & \multicolumn{1}{l}{83.86} & \multicolumn{1}{l}{59.45} & \multicolumn{1}{l}{73.91} & \multicolumn{1}{l|}{\textbf{65.66}} & \multicolumn{1}{l}{81.35} & \multicolumn{1}{l}{78.92} \\ 
\cmidrule[0.7pt]{2-18}
 & \multicolumn{1}{l|}{*Remote Sensing VQA Methods} & \multicolumn{1}{l}{} & \multicolumn{1}{l}{} & \multicolumn{1}{l|}{} & \multicolumn{1}{l}{} & \multicolumn{1}{l}{} & \multicolumn{1}{l}{} & \multicolumn{1}{l}{} & \multicolumn{1}{l}{} & \multicolumn{1}{l|}{} & \multicolumn{1}{l}{} & \multicolumn{1}{l}{} & \multicolumn{1}{l}{} & \multicolumn{1}{l}{} & \multicolumn{1}{l|}{} & \multicolumn{1}{l}{} & \multicolumn{1}{l}{} \\
& RSVQA (TGRS, 2020)\cite{lobry2020rsvqa} & 85.57 & 96.98 & 98.47 & 46.59 & 79.12 & 83.64 & 95.26 & 95.88 & 97.64 & 75.71 & 86.73 & 53.84& 66.20 & 59.85 & 82.39 & 80.11 \\
& RSIVQA (TGRS, 2021)\cite{Zheng2021Mutual} & 88.90 & 96.61 & 98.27 & 41.49 & 78.00 & 83.73 & 94.46 & 93.23 & 95.59 & 71.06 & 82.76 & 47.82 & 63.56 & 59.11 & 80.71 & 78.19 \\
& FEH (TGRS, 2022)\cite{yuan2022easy} & 85.86 & 98.04 & 98.52 & 56.40 & 84.39 & 89.33 & 96.05 & 96.85 & 98.78 & 83.85 & 91.52 & 77.34 & 88.42 & 60.81 & 88.12 & 86.15 \\
& MQVQA (TGRS, 2023)\cite{zhang2023multi} & 90.12 & 97.77 & 98.94 & 52.16 & 82.44 & 88.35 & 97.41 & 97.42 & 98.38 & 80.05 & 89.56 & 62.43 & 73.08 & 61.43 & 85.67  &  83.54 \\
& EarthVQA (AAAI, 2024)\cite{wang2024earthvqa} & 86.62 & 98.00 & \textbf{99.74} & 66.24 & 87.59 & 92.26 & 97.59 & 96.81 & 98.76 & 91.18 & 94.87 & 84.65 & 92.16 & 62.11 & 90.81 & 89.18 \\
& SGA (IGARSS, 2024)\cite{tosato2024segmentation} & 95.71 & 97.56 & 99.09 & 69.85 & 89.03 & 95.13 & 96.93 & 98.83 & 99.41 & 89.03 & 94.82 & 73.53 & 84.14 & 58.95 & 90.51 & 88.72 \\
\cmidrule[0.7pt]{2-18}
\rowcolor[HTML]{C8F3FF} 
\cellcolor{white} & \cellcolor[HTML]{C8F3FF}\textbf{CMFPF (Ours)} & \textbf{96.68} & \textbf{98.30} & 99.49 & \textbf{80.85} & \textbf{92.43} & \textbf{96.70} & \textbf{99.26} & \textbf{99.51} & \textbf{99.45} & \textbf{92.22} & \textbf{96.59} & \textbf{87.52} & \textbf{92.87} & 62.34 & \textbf{93.85} & \textbf{92.44} \\ 
\midrule[0.7pt]
\multirow{14}{*}{\rotatebox{90}{RS-CMQA-B Test Set}} & \multicolumn{1}{l|}{*General VQA Methods} & \multicolumn{1}{l}{} & \multicolumn{1}{l}{} & \multicolumn{1}{l|}{} & \multicolumn{1}{l}{} & \multicolumn{1}{l}{} & \multicolumn{1}{l}{} & \multicolumn{1}{l}{} & \multicolumn{1}{l}{} & \multicolumn{1}{l|}{} & \multicolumn{1}{l}{} & \multicolumn{1}{l}{} & \multicolumn{1}{l}{} & \multicolumn{1}{l}{} & \multicolumn{1}{l|}{} & \multicolumn{1}{l}{} & \multicolumn{1}{l}{} \\
& SAN (CVPR, 2016)\cite{yang2016stacked} & 78.84 & 93.72 & 96.84 & 23.42 & 58.20 & 53.29 & 76.44 & 78.87 & 87.29 & 47.07 & 56.64 & 42.70 & 54.06 & 51.61 & 64.17 & 64.21 \\
& MAC (ICLR, 2018)\cite{hudson2018compositional} & 76.46 & 92.70 & 96.61 & 24.52 & 63.12 & 56.27 & 82.19 & 78.47 & 88.44 & 51.44 & 72.41 & 44.66 & 59.86 & 57.05 & 67.42 & 67.44 \\
& MCAN (CVPR, 2019)\cite{yu2019deep} &  64.74 & 87.51 & 94.04 & 20.79 & 52.53 & 32.52 & 57.47 & 69.03 & 79.45 & 24.80 & 47.42 & 40.46 & 50.37 & 51.08 & 55.12 & 55.16 \\
& DVQA (NeurIPS, 2021)\cite{wen2021debiased} & 79.26 & 92.65 & 95.00 & 20.81 & 62.93 & 62.08 & 81.24 & 78.56 & 85.17 & 52.52 & 72.98 & 42.82 & 49.97 & \textbf{59.15} & 66.77 & 66.78 \\
& BLIP-2 (ICML, 2023)\cite{li2023blip}  & \multicolumn{1}{l}{79.43} & \multicolumn{1}{l}{91.44} & \multicolumn{1}{l|}{91.44} & \multicolumn{1}{l}{20.69} & \multicolumn{1}{l}{54.58} & \multicolumn{1}{l}{57.39} & \multicolumn{1}{l}{80.58} & \multicolumn{1}{l}{77.85} & \multicolumn{1}{l|}{86.61} & \multicolumn{1}{l}{47.59} & \multicolumn{1}{l}{72.43} & \multicolumn{1}{l}{43.84} & \multicolumn{1}{l}{54.74} & \multicolumn{1}{l|}{56.33} & \multicolumn{1}{l}{65.32} & \multicolumn{1}{l}{65.35} \\ \cmidrule[0.7pt]{2-18}
& \multicolumn{1}{l|}{*Remote Sensing VQA Methods} & \multicolumn{1}{l}{} & \multicolumn{1}{l}{} & \multicolumn{1}{l|}{} & \multicolumn{1}{l}{} & \multicolumn{1}{l}{} & \multicolumn{1}{l}{} & \multicolumn{1}{l}{} & \multicolumn{1}{l}{} & \multicolumn{1}{l|}{} & \multicolumn{1}{l}{} & \multicolumn{1}{l}{} & \multicolumn{1}{l}{} & \multicolumn{1}{l}{} & \multicolumn{1}{l|}{} & \multicolumn{1}{l}{} & \multicolumn{1}{l}{} \\
& RSVQA (TGRS, 2020)\cite{lobry2020rsvqa} & 70.55 & 91.38 & 96.06 & 29.09 & 64.04 & 49.49 & 77.34 & 67.33 & 81.31 & 48.58 & 69.65 & 39.57 & 55.98 & 50.69 & 63.62 & 63.65 \\
& RSIVQA (TGRS, 2021)\cite{Zheng2021Mutual} & 62.48 & 90.23 & 96.19 & 25.21 & 63.92 & 52.42 & 77.62 & 63.05 & 78.51 & 49.28 & 68.75 & 33.55 & 49.90 & 50.33 & 61.53 & 61.53 \\
& FEH (TGRS, 2022)\cite{yuan2022easy} & 75.01 & 90.04 & 95.42 & 24.44 & 64.82 & 55.49 & 80.35 & 72.37 & 85.14 & 51.70 & 73.62 & 43.26 & 58.18 & 52.87 & 65.88 & 65.91 \\
& MQVQA (TGRS, 2023)\cite{zhang2023multi} & 65.85 & 92.99 & 95.85 & 27.31 & 67.92 & 53.37 & 82.08 & 63.75 & 81.13 & 51.97 & 74.20 & 36.08 & 54.81 & 54.58 & 64.42 & 64.42 \\
& EarthVQA (AAAI, 2024)\cite{wang2024earthvqa} & 83.12 & 91.63 & 95.87 & 23.34 & 62.34 & 65.79 & 82.52 & 86.16 & 91.41 & \textbf{62.48} & 74.95 & 45.67 & 63.82 & 51.04 & 69.97 & 70.01 \\
& SGA (IGARSS, 2024)\cite{tosato2024segmentation} & 85.39 & 92.16 & 95.25 & 25.00 & 61.67 & 68.73 & 83.43 & 88.16 & 93.14 & 58.23 & 77.09 & 47.44 & 63.45 & 53.64 & 70.88 & 70.91 \\
\cmidrule[0.7pt]{2-18}
\rowcolor[HTML]{C8F3FF} 
\cellcolor{white} & \cellcolor[HTML]{C8F3FF}\textbf{CMFPF (Ours)} & \textbf{87.64} & \textbf{92.49} & \textbf{97.00} & \textbf{33.56} & \textbf{67.03} & \textbf{83.31} & \textbf{90.01} & \textbf{91.72} & \textbf{94.74} & 62.36 & \textbf{80.35} & \textbf{50.94} & \textbf{67.87} & 58.31 & \textbf{75.51} & \textbf{75.52}\\ 
\midrule[0.7pt]
\multirow{14}{*}{\rotatebox{90}{Real-RSCM Test Set}} & \multicolumn{1}{l|}{*General VQA Methods} & \multicolumn{1}{l}{} & \multicolumn{1}{l}{} & \multicolumn{1}{l|}{} & \multicolumn{1}{l}{} & \multicolumn{1}{l}{} & \multicolumn{1}{l}{} & \multicolumn{1}{l}{} & \multicolumn{1}{l}{} & \multicolumn{1}{l|}{} & \multicolumn{1}{l}{} & \multicolumn{1}{l}{} & \multicolumn{1}{l}{} & \multicolumn{1}{l}{} & \multicolumn{1}{l|}{} & \multicolumn{1}{l}{} & \multicolumn{1}{l}{} \\
& SAN (CVPR, 2016)\cite{yang2016stacked} & 85.71 & 96.04 & 98.46 & 37.15 & 64.94 & 30.78 & 66.74 & 90.23 & 93.56 & 36.74 & 64.45 & 39.78 & 58.97 & 52.89 & 68.44 & 65.46 \\
& MAC (ICLR, 2018)\cite{hudson2018compositional} & 86.63 & 98.18 & 98.53 & 49.78 & 83.42 & 62.32 & 86.37 & 93.74 & 95.11 & 80.84 & 90.61 & 77.63 & 87.23 & 55.34 & 84.33 & 81.84 \\
& MCAN (CVPR, 2019)\cite{yu2019deep} & 83.81 & 96.05 & 98.29 & 29.46 & 67.98 & 33.16 & 75.77 & 89.20 & 92.90 & 31.32 & 71.82 & 71.42 & 62.40 & 54.62 & 71.54 & 68.44 \\
& DVQA (NeurIPS, 2021)\cite{wen2021debiased} & 90.20 & 97.20 & 98.40 & 44.57 & 77.37 & 74.10 & 87.95 & 92.81 & 97.53 & 78.13 & 87.64 & 71.59 & 85.90 & 57.17 & 83.82 & 81.47 \\
& BLIP-2 (ICML, 2023)\cite{li2023blip}  & \multicolumn{1}{l}{90.23} & \multicolumn{1}{l}{96.42} & \multicolumn{1}{l|}{99.11} & \multicolumn{1}{l}{49.44} & \multicolumn{1}{l}{73.94} & \multicolumn{1}{l}{47.85} & \multicolumn{1}{l}{77.40} & \multicolumn{1}{l}{87.55} & \multicolumn{1}{l|}{90.33} & \multicolumn{1}{l}{49.68} & \multicolumn{1}{l}{60.83} & \multicolumn{1}{l}{59.70} & \multicolumn{1}{l}{65.58} & \multicolumn{1}{l|}{57.36} & \multicolumn{1}{l}{73.89} & \multicolumn{1}{l}{71.82} \\ 
\cmidrule[0.7pt]{2-18}
& \multicolumn{1}{l|}{*Remote Sensing VQA Methods} & \multicolumn{1}{l}{} & \multicolumn{1}{l}{} & \multicolumn{1}{l|}{} & \multicolumn{1}{l}{} & \multicolumn{1}{l}{} & \multicolumn{1}{l}{} & \multicolumn{1}{l}{} & \multicolumn{1}{l}{} & \multicolumn{1}{l|}{} & \multicolumn{1}{l}{} & \multicolumn{1}{l}{} & \multicolumn{1}{l}{} & \multicolumn{1}{l}{} & \multicolumn{1}{l|}{} & \multicolumn{1}{l}{} & \multicolumn{1}{l}{} \\
& RSVQA (TGRS, 2020)\cite{lobry2020rsvqa} & 88.66 & 96.82 & 98.03 & 50.24 & 78.74 & 61.44 & 84.72 & 92.18 & 95.57 & 72.01 & 86.17 & 43.26 & 63.13 & 53.58 & 78.64 & 76.04 \\
& RSIVQA (TGRS, 2021)\cite{Zheng2021Mutual} & 83.38 & 94.73 & 98.38 & 30.42 & 67.87 & 26.00 & 68.29 & 91.58 & 94.28 & 40.47 & 58.79 & 32.02 & 49.68 & 55.03 & 66.69 & 63.64 \\
& FEH (TGRS, 2022)\cite{yuan2022easy} & 93.03 & 97.26 & 98.54 & 58.58 & 85.02 & 77.97 & 92.43 & 93.02 & 96.81 & 84.20 & 92.04 & 61.66 & 80.56 & 55.91 & 85.67 & 83.36 \\
& MQVQA (TGRS, 2023)\cite{zhang2023multi} & 89.73 & 96.28 & 98.10 & 61.40 & 81.06 & 81.20 & 91.35 & 95.35 & 97.21 & 61.53 & 79.59 & 44.55 & 63.26 & 58.14 & 80.62 & 78.48\\
& EarthVQA (AAAI, 2024)\cite{wang2024earthvqa} & 86.48 & 94.27 & 96.86 & 56.12 & 81.00 & 61.65 & 82.65 & 92.85 & 94.39 & 85.02 & 90.47 & 78.89 & 87.48 & 52.29 & 83.63 & 81.46 \\
& SGA (IGARSS, 2024)\cite{tosato2024segmentation} & 89.34 & 96.26 & 96.03 & 57.48 & 83.32 & 63.70 & 84.02 & 93.92 & 97.28 & 84.53 & 92.57 & 79.30 & 87.20 & 57.62 & 85.14 & 83.04\\
\cmidrule[0.7pt]{2-18}
\rowcolor[HTML]{C8F3FF} 
\cellcolor{white} & \cellcolor[HTML]{C8F3FF}\textbf{CMFPF (Ours)} &  \textbf{94.47} & \textbf{97.66} & \textbf{99.92} & \textbf{84.61} & \textbf{93.83} & \textbf{93.10} & \textbf{95.89} & \textbf{98.74} & \textbf{99.26} & \textbf{91.64} & \textbf{96.85} & \textbf{85.36} & \textbf{90.64} & \textbf{59.59} & \textbf{92.86} & \textbf{91.54} \\ 
\midrule[0.7pt]
\end{tabular}
}
\vspace{-7pt}
\label{tab:cmqa}
\end{table*}

\begin{table*}[t]
\centering
\caption{
Evaluation with state-of-the-art methods on the RS-TQA test set and RS-TQA-B test set, with best metrics highlighted in \textbf{bold}.
}
\vspace{-6pt}
\resizebox{1\textwidth}{!}{
\renewcommand{\arraystretch}{1.1}
\large 
\begin{tabular}{c|c|cccc|cccccc|ccccc|cc}
\midrule[0.9pt]
 &  & \multicolumn{4}{c|}{Basic Questions} & \multicolumn{6}{c|}{Independent Questions} & \multicolumn{5}{c|}{Related Questions} &  &  \\
 & \multirow{-2}{*}{Method} & Q1 & Q2 & Q3 & Q4 & Q5 & Q6 & Q7 & Q8 & Q9 & Q10 & Q11 & Q12 & Q13 & Q14 & Q15 &\multirow{-2}{*}{OA} & \multirow{-2}{*}{AA} \\ \midrule[0.9pt]
\multirow{14}{*}{\rotatebox{90}{RS-TQA Test Set}}  & \multicolumn{1}{l|}{*General VQA Methods} & \multicolumn{1}{l}{} & \multicolumn{1}{l}{} & \multicolumn{1}{l}{}& \multicolumn{1}{l|}{} & \multicolumn{1}{l}{} & \multicolumn{1}{l}{} & \multicolumn{1}{l}{} & \multicolumn{1}{l}{} & \multicolumn{1}{l}{} & \multicolumn{1}{l|}{} & \multicolumn{1}{l}{} & \multicolumn{1}{l}{} & \multicolumn{1}{l}{} & \multicolumn{1}{l}{} & \multicolumn{1}{l|}{} & \multicolumn{1}{l}{} & \multicolumn{1}{l}{} \\
 & SAN (CVPR, 2016)\cite{yang2016stacked} & 93.07 & 98.98 & 98.81 & 99.54 & 69.22 & 88.69 & 89.13 & 96.05 & 97.26 & 98.45 & 70.80 & 84.08 & 62.99 & 75.72 & 59.83 & 89.22 & 85.51 \\
 & MAC (ICLR, 2018)\cite{hudson2018compositional} & 90.36 & 97.09 & 98.37 & 98.79 & 71.04 & 90.89 & 85.66 & 95.00 & 96.64 & 97.95 & 78.48 & 87.91 & 73.75 & 83.49 & 64.53 & 90.31 & 87.33\\
 & MCAN (CVPR, 2019)\cite{yu2019deep} & 77.74& 94.54& 96.96& 98.29& 68.61& 87.44& 69.92& 82.05& 94.59& 97.28& 72.58& 85.91& 46.13& 61.85& 55.96 & 84.08 & 79.33 \\
 & DVQA (NeurIPS, 2021)\cite{wen2021debiased} & 90.23& 97.64& 98.07& 99.34& 71.16& 89.36& 86.64& 94.15& 97.52& 98.69& 78.44& 87.22& 51.83& 61.42& \textbf{67.44} & 88.31 & 84.61 \\
 & BLIP-2 (ICML, 2023)\cite{li2023blip}  & \multicolumn{1}{l}{89.95} & \multicolumn{1}{l}{97.30} & \multicolumn{1}{l}{97.91}& \multicolumn{1}{l|}{98.62} & \multicolumn{1}{l}{69.43} & \multicolumn{1}{l}{88.27} & \multicolumn{1}{l}{81.06} & \multicolumn{1}{l}{92.63} & \multicolumn{1}{l}{95.87} & \multicolumn{1}{l|}{97.56} & \multicolumn{1}{l}{72.15} & \multicolumn{1}{l}{83.96} & \multicolumn{1}{l}{50.65} & \multicolumn{1}{l}{60.86} & \multicolumn{1}{l|}{63.38} & \multicolumn{1}{l}{86.82} & \multicolumn{1}{l}{82.64}  \\ 
\cmidrule[0.7pt]{2-19}
 & \multicolumn{1}{l|}{*Remote Sensing VQA Methods} & \multicolumn{1}{l}{} & \multicolumn{1}{l}{}& \multicolumn{1}{l}{} & \multicolumn{1}{l|}{} & \multicolumn{1}{l}{} & \multicolumn{1}{l}{} & \multicolumn{1}{l}{} & \multicolumn{1}{l}{} & \multicolumn{1}{l}{} & \multicolumn{1}{l|}{} & \multicolumn{1}{l}{} & \multicolumn{1}{l}{} & \multicolumn{1}{l}{} & \multicolumn{1}{l}{} & \multicolumn{1}{l|}{} & \multicolumn{1}{l}{} & \multicolumn{1}{l}{}  \\
 & RSVQA (TGRS, 2020)\cite{lobry2020rsvqa} & 89.97 & 97.79 & 97.91 & 99.19 & 69.0 & 89.72 & 84.55 & 95.37 & 96.0 & 96.87 & 74.65 & 86.91 & 52.42 & 65.25 & 59.77 & 87.76 & 83.69\\
 & RSIVQA (TGRS, 2021)\cite{Zheng2021Mutual} & 88.46 & 94.45 & 97.17 & 98.81 & 71.83 & 89.35 & 64.94 & 75.35 & 96.31 & 97.62 & 79.51 & 90.26 & 49.66 & 66.68 & 58.37 & 86.03 & 81.25 \\
 & FEH (TGRS, 2022)\cite{yuan2022easy} & 90.69 & 98.68 & 98.14 & 99.32 & 74.53 & 89.98 & 87.19 & 95.77 & 98.12 & 97.89 & 79.00 & 89.43 & 67.60 & 82.07 & 62.84 & 90.54 & 87.42\\
 & MQVQA (TGRS, 2023)\cite{zhang2023multi} & 93.16 & 98.75 & 98.82 & 98.68 & 73.14 & 91.15 & 87.60 & 95.72 & 96.30 & 97.32 & 74.76 & 87.39 & 61.19 & 75.73 & 58.64 & 89.62 & 85.89 \\
 & EarthVQA (AAAI, 2024)\cite{wang2024earthvqa} & 96.92 & 98.99 & 97.87 & 99.40 & 79.42 & 92.39 & 92.44 & 95.18 & 99.14 & 99.43 & 86.06 & 90.91 & 57.35 & 72.23 & 59.99 & 91.33 & 87.85 \\
 & SGA (IGARSS, 2024)\cite{tosato2024segmentation} & 96.34 & 99.59 & 97.92 & 99.30 & 81.07 & 92.17 & 92.79 & 96.24 & 98.55 & \textbf{99.61} & 85.31 & 92.89 & 60.43 & 71.70 & 60.93 & 91.64 & 88.32\\
\cmidrule[0.7pt]{2-19}
\rowcolor[HTML]{C8F3FF} 
 \cellcolor{white} & \textbf{CMFPF (Ours)} & \textbf{97.08} & \textbf{99.75} & \textbf{99.53} & \textbf{99.28} & \textbf{86.89} & \textbf{94.64} & \textbf{93.72} & \textbf{98.06} & \textbf{99.25} & 99.41 & \textbf{90.84} & \textbf{94.95} & \textbf{77.71} & \textbf{84.88} & 65.57 & \textbf{94.43} & \textbf{92.10} \\ 
 \midrule[0.7pt]
 \multirow{14}{*}{\rotatebox{90}{RS-TQA-B Test Set}}  & \multicolumn{1}{l|}{*General VQA Methods} & \multicolumn{1}{l}{} & \multicolumn{1}{l}{} & \multicolumn{1}{l}{}& \multicolumn{1}{l|}{} & \multicolumn{1}{l}{} & \multicolumn{1}{l}{} & \multicolumn{1}{l}{} & \multicolumn{1}{l}{} & \multicolumn{1}{l}{} & \multicolumn{1}{l|}{} & \multicolumn{1}{l}{} & \multicolumn{1}{l}{} & \multicolumn{1}{l}{} & \multicolumn{1}{l}{} & \multicolumn{1}{l|}{} & \multicolumn{1}{l}{} & \multicolumn{1}{l}{} \\
& SAN (CVPR, 2016)\cite{yang2016stacked} & 86.30 & 95.58 & 96.18 & 98.40 & 54.62 & 79.40 & 67.60 & 87.25 & 84.74 & 92.43 & 50.86 & 68.93 & 47.65 & 62.15 & 57.40 & 75.39 & 75.30 \\
& MAC (ICLR, 2018)\cite{hudson2018compositional} & 83.46 & 91.67 & 95.89 & 97.31 & 54.06 & 80.48 & 66.79 & 86.45 & 83.57 & 90.06 & 55.39 & 74.76 & 51.23 & 66.24 & 59.56 & 75.89 & 75.79 \\
& MCAN (CVPR, 2019)\cite{yu2019deep} & 84.40 & 95.72 & 94.78 & 97.22 & 47.92 & 66.95 & 52.98 & 77.77 & 77.47 & 87.50 & 46.50 & 52.35 & 39.49 & 50.18 & 53.63 & 68.40 & 68.32 \\
& DVQA (NeurIPS, 2021)\cite{wen2021debiased} & 83.62 & 91.16 & 95.77 & 97.51 & 54.09 & 79.39 & 65.47 & 85.33 & 85.29 & 90.42 & 53.62 & 71.41 & 42.84 & 50.74 & 60.69 & 73.93 & 73.82 \\
& BLIP-2 (ICML, 2023)\cite{li2023blip}  & \multicolumn{1}{l}{84.56} & \multicolumn{1}{l}{93.48} & \multicolumn{1}{l}{92.50}& \multicolumn{1}{l|}{96.07} & \multicolumn{1}{l}{49.84} & \multicolumn{1}{l}{73.18} & \multicolumn{1}{l}{68.71} & \multicolumn{1}{l}{82.94} & \multicolumn{1}{l}{79.65} & \multicolumn{1}{l|}{88.50} & \multicolumn{1}{l}{50.22} & \multicolumn{1}{l}{73.83} & \multicolumn{1}{l}{45.58} & \multicolumn{1}{l}{52.03} & \multicolumn{1}{l|}{54.96} & \multicolumn{1}{l}{72.52} & \multicolumn{1}{l}{72.40}  \\ 
\cmidrule[0.7pt]{2-19}
& \multicolumn{1}{l|}{*Remote Sensing VQA Methods} & \multicolumn{1}{l}{} & \multicolumn{1}{l}{}& \multicolumn{1}{l}{} & \multicolumn{1}{l|}{} & \multicolumn{1}{l}{} & \multicolumn{1}{l}{} & \multicolumn{1}{l}{} & \multicolumn{1}{l}{} & \multicolumn{1}{l}{} & \multicolumn{1}{l|}{} & \multicolumn{1}{l}{} & \multicolumn{1}{l}{} & \multicolumn{1}{l}{} & \multicolumn{1}{l}{} & \multicolumn{1}{l|}{} & \multicolumn{1}{l}{} & \multicolumn{1}{l}{}  \\
& RSVQA (TGRS, 2020)\cite{lobry2020rsvqa} & 77.47 & 90.44 & 92.52 & 96.84 & 50.39 & 77.54 & 55.12 & 82.54 & 77.22 & 86.73 & 50.76 & 74.44 & 44.02 & 60.87 & 53.39 & 71.45 & 71.35 \\
& RSIVQA (TGRS, 2021)\cite{Zheng2021Mutual} & 79.74 & 96.14 & 92.88 & 96.13 & 56.25 & 75.56 & 58.07 & 79.48 & 77.98 & 86.28 & 47.10 & 65.70 & 38.54 & 48.44 & 53.38 & 70.21 & 70.11 \\
& FEH (TGRS, 2022)\cite{yuan2022easy} & 83.02 & 93.07 & 95.35 & 98.17 & 54.74 & 76.97 & 64.79 & 85.73 & 83.61 & 89.48 & 55.99 & 77.17 & 46.67 & 68.32 & 56.98 & 75.44 & 75.34\\
& MQVQA (TGRS, 2023)\cite{zhang2023multi} & 81.26 & 95.64 & 96.20 & 95.92 & 57.59 & 80.76 & 64.97 & 85.80 & 79.89 & 88.30 & 53.17 & 77.38 & 41.57 & 58.97 & 54.79 & 74.25 & 74.15 \\
& EarthVQA (AAAI, 2024)\cite{wang2024earthvqa} & 94.52 & 98.65 & 94.10 & \textbf{98.54} & 53.23 & 76.66 & 81.71 & 90.85 & 94.35 & 96.53 & 55.61 & 73.76 & 46.81 & 64.78 & 55.03 & 78.47 & 78.34 \\
& SGA (IGARSS, 2024)\cite{tosato2024segmentation} & 94.49 & 99.08 & 93.16 & 97.53 & 55.75 & 77.32 & \textbf{82.66} & 91.92 & 93.97 & 96.76 & 56.18 & 78.60 & 51.63 & 64.35 & 54.54 & 79.32 & 79.20\\
\cmidrule[0.7pt]{2-19}
\rowcolor[HTML]{C8F3FF} 
\cellcolor{white} & \textbf{CMFPF (Ours)} & \textbf{95.32} & \textbf{99.25} & \textbf{96.54} & 98.15 & \textbf{62.16} & \textbf{81.67} & 82.50 & \textbf{93.45} & \textbf{95.98} & \textbf{97.91} & \textbf{66.41} & \textbf{82.43} & \textbf{55.29} & \textbf{71.36} & \textbf{60.69} & \textbf{82.72}& \textbf{82.61} \\ \midrule[0.7pt]
\end{tabular}
}
\vspace{-7pt}
\label{tab:tqa}
\end{table*}

\begin{figure*}[!t]
    \centering
    \includegraphics[width=1\linewidth]{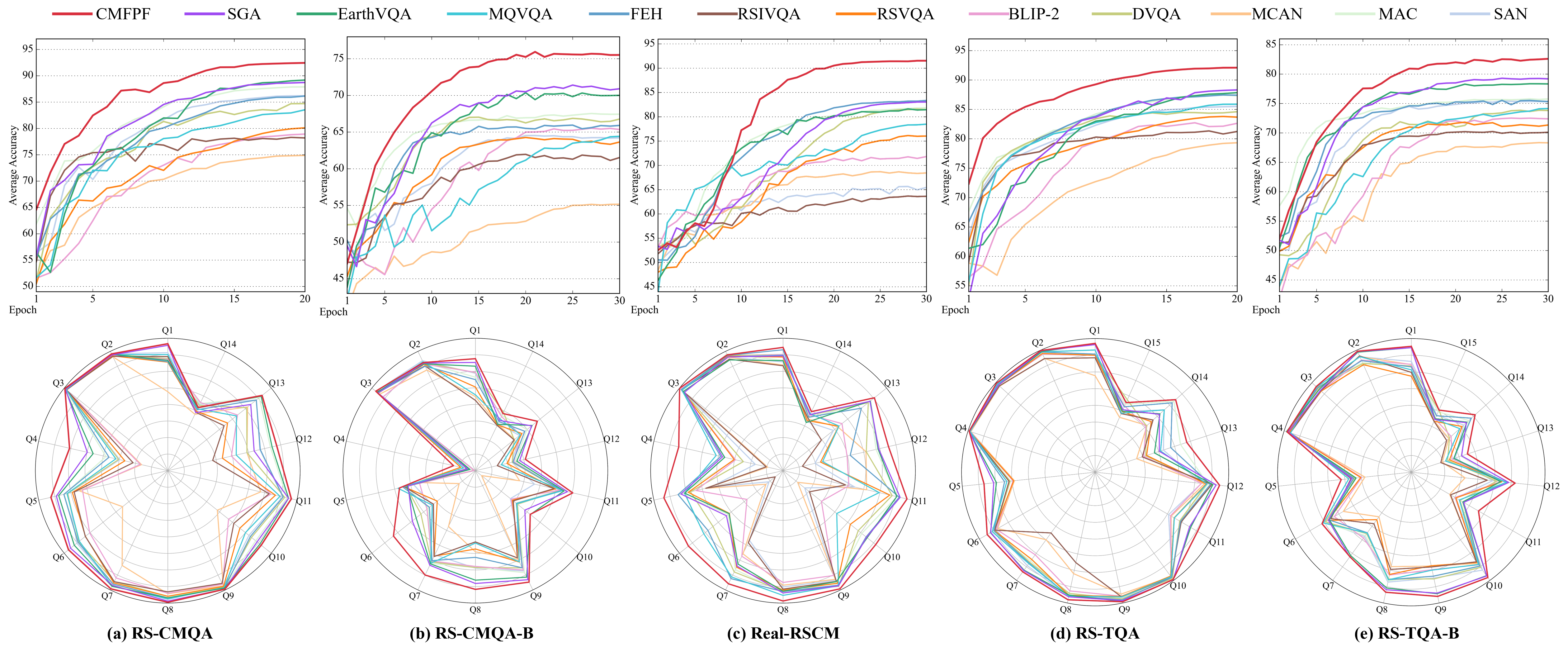}
    \vspace{-18pt}
    \caption{The overall accuracy of the models per epoch on the validation set of the five datasets, as well as the accuracy coverage across different problem categories on the test set of the five datasets. CMFPF demonstrates a stable and significant performance advantage.
    }
    \vspace{-2pt}
    \label{fig:radar}
\end{figure*}

\vspace{-3pt}
\subsection{Comparative experiments}
\noindent \textbf{Baseline Comparison.}  
Eleven advanced models were selected as baselines. These include SAN\cite{yang2016stacked}, MAC\cite{hudson2018compositional}, MCAN\cite{yu2019deep}, DVQA\cite{wen2021debiased} and BLIP-2-2.7B\cite{li2023blip} as classic general question-answering models, and RSVQA\cite{lobry2020rsvqa}, RSIVQA\cite{Zheng2021Mutual}, FEH\cite{yuan2022easy}, MQVQA\cite{zhang2023multi}, SGA\cite{tosato2024segmentation} and EarthVQA\cite{wang2024earthvqa} specifically designed for remote sensing tasks. The experimental results, as summarized in Table \ref{tab:cmqa}, indicate that most baseline models perform well on fundamental questions such as Q2 and Q3. However, accuracy declines considerably for tampering-related questions (Q1) as well as independent and related questions, highlighting the complexity and challenges of the RSCMQA task. SAN, MCAN, and RSIVQA attempted post-fusion feature enhancement, yet yielded marginal improvements. Despite its large parameter count, BLIP-2-2.7B fails to exhibit a performance advantage, suggesting that merely increasing model capacity offers limited benefits without targeted feature extraction for tampered regions. In contrast, MAC and DVQA improve predictions through specialized network architectures and cross-modal feature alignment. FEH employs a difficulty-aware loss function to facilitate learning of challenging questions. Additionally, EarthVQA and SGA leverage semantic segmentation prompts for question answering, demonstrating relatively strong performance. However, semantic segmentation alone does not effectively distinguish between source and tampered regions, thus failing to provide clear guidance for the question-answering model.

The proposed CMFPF achieves state-of-the-art performance across the RS-CMQA, RS-CMQA-B, and Real-RSCM datasets. Specifically, on the RS-CMQA dataset, CMFPF attains the best accuracy in 12 out of 14 question categories, surpassing the second-best model by 3.04\% in OA and 3.26\% in AA. RS-CMQA is a large but imbalanced dataset, allowing models to acquire substantial domain knowledge. However, this imbalance may introduce bias in question-answering models, limiting their ability to fully reflect performance disparities.
All methods perform worse on the RS-CMQA-B dataset than on RS-CMQA, likely due to the smaller and more balanced nature of RS-CMQA-B, which prevents models from exploiting data distribution biases. CMFPF demonstrates a more pronounced advantage on RS-CMQA-B, achieving the highest accuracy in 12 out of 14 question categories, with OA and AA improvements of 4.63\% and 4.61\%, respectively, over the second-best model.
Real-RSCM is a manually annotated high-quality dataset, where most tampered objects are visually imperceptible. This makes it a more realistic benchmark for assessing the true potential of models in real-world tampering scenarios. Most baseline models experience severe performance degradation on Real-RSCM compared to RS-CMQA—for instance, SAN exhibits a 19.78\% drop in OA, while RSIVQA declines by 14.02\%. In contrast, CMFPF maintains robustness, with only a 1.09\% decrease. CMFPF outperforms the second-best model on Real-RSCM by 7.72\% in OA and 8.50\% in AA, underscoring its superior capability in handling real-world tampering cases.
It is worth noting that all models exhibit low accuracy on Q14, which assesses whether objects subjected to copy-move tampering have been rotated. This task requires precise spatial localization of both source and tampered regions, an area where current models still face notable limitations.
Figure \ref{fig:radar} illustrates the overall validation accuracy curves throughout training and the per-category accuracy radar charts on the test set for various methods. The results highlight the distinct and consistent performance advantage demonstrated by CMFPF.

\begin{table}[!t]
\centering
\caption{Comparison of the efficiency of different methods.}
\label{tab:flops}
\resizebox{1\linewidth}{!}{
\begin{tabular}{cccc}
\midrule[0.8pt]
Method   & Params (M) & FLOPs (G) & Throughput (case/second) \\ \midrule[0.8pt]
MAC\cite{hudson2018compositional}      & 14.29      & 41.82     & 141.37                   \\
MCAN\cite{yu2019deep}     & 15.18      & 149.36    & 133.28                   \\
RSVQA\cite{lobry2020rsvqa}    & 15.34      & 85.62     & 135.63                   \\
DVQA\cite{wen2021debiased}     & 16.85      & 120.89    & 132.95                   \\
MQVQA\cite{zhang2023multi}    & 41.26      & 38.36     & 81.38                    \\
RSIVQA\cite{Zheng2021Mutual}   & 81.01      & 302.99    & 68.51                    \\
SAN\cite{yang2016stacked}      & 81.57      & 78.54     & 110.40                   \\
FEH\cite{yuan2022easy}      & 121.58     & 146.28    & 114.25                   \\
EarthVQA\cite{wang2024earthvqa} & 200.08     & 200.01    & 89.52                    \\
SGA\cite{tosato2024segmentation}      & 232.64     & 156.08    & 91.77                    \\ \midrule[0.8pt]
\textbf{CMFPF}    & 54.41      & 197.10    & 111.92                   \\ \midrule[0.8pt]
\end{tabular}
}
\end{table}

\noindent \textbf{Efficiency Analysis.} 
Params denote the total trainable parameters, reflecting the memory footprint of a model. FLOPs measure the theoretical computational cost of a single forward pass. Throughput refers to the number of test samples processed per second during inference, serving as a practical efficiency indicator. Importantly, inference speed is not linearly determined by parameter size or FLOPs; it is shaped by factors such as parameter distribution, computational graph design, pipeline bottlenecks, code-level optimization, and hardware acceleration. For example, modern GPU architectures often favor Transformer-based models with faster inference despite comparable or larger parameter counts. In this study, all baseline models were faithfully reproduced without any special optimization for inference. Similarly, our proposed CMFPF framework was evaluated under identical conditions. Therefore, the throughput values reported in Table \ref{tab:flops} should be interpreted as reference measurements under unified experimental settings, rather than the fastest achievable inference speeds. This ensures fair comparison across methods and provides a more realistic view of efficiency trade-offs.

From the results, our CMFPF model has a parameter count only slightly higher than lightweight baselines such as MAC\cite{hudson2018compositional}, MCAN\cite{yu2019deep}, RSVQA\cite{lobry2020rsvqa}, DVQA\cite{wen2021debiased}, and MQVQA\cite{zhang2023multi}. In contrast, models incorporating complex fusion mechanisms (RSIVQA\cite{Zheng2021Mutual}, SAN\cite{yang2016stacked}), difficulty-aware modules (FEH\cite{yuan2022easy}), or global segmentation mask modeling (EarthVQA\cite{wang2024earthvqa}, SGA\cite{tosato2024segmentation}) exhibit significantly larger parameter sizes. However, as shown in Table \ref{tab:flops}, their increased scale does not translate into improved accuracy, as the core challenge of this task lies in modeling the correlation and discrepancy between tampered and original regions. Without targeted cross-modal fusion, global feature extraction and mask injection introduce additional computational overhead without enhancing performance.

\begin{table}[!t]
\caption{Experimental performance of different modules on the CMFPF architecture.}
\label{tab:module_analysis}
\centering
\large
\renewcommand{\arraystretch}{1.1}
\resizebox{1.0\linewidth}{!}{
\begin{tabular}{ccc|cc|cc|cc}
\midrule[1.3pt]
\multirow{2}{*}{\begin{tabular}[c]{@{}c@{}}Tampering\\ Head\end{tabular}} & \multirow{2}{*}{\begin{tabular}[c]{@{}c@{}}Visual\\ Head\end{tabular}} & \multirow{2}{*}{\begin{tabular}[c]{@{}c@{}}Text\\ Head\end{tabular}} & \multicolumn{2}{c|}{RS-CMQA} & \multicolumn{2}{c|}{RS-CMQA-B} & \multicolumn{2}{c}{Real-RSCM}\\
 &  &  & OA & AA & OA & AA & OA & AA   \\ 
\midrule[1.3pt]
Swin & Res-152 & LSTM & 80.88 & 78.70 & 66.94 & 67.00 & 82.80 & 81.64 \\
Swin & Res-152 & BERT & 88.50 & 86.45 & 73.34 & 73.41 & 90.66 & 88.90 \\
Swin & ViT-B & BERT & 91.57 & 89.86 & 74.02 & 74.04 & 91.66 & 90.20 \\
U-Net & Res-152 & LSTM & 84.43 & 82.39 & 68.33 & 68.39 & 83.17 & 82.17 \\
U-Net & Res-152 & BERT & 93.76 & 92.27 & 73.88 & 73.89 & 92.26 & 90.73 \\
\textbf{U-Net} & \textbf{ViT-B} & \textbf{BERT} & \textbf{93.85} & \textbf{92.44} & \textbf{75.51} & \textbf{75.52} & \textbf{92.86} & \textbf{91.54}   \\ 
\midrule[1.3pt]
\end{tabular}
}
\end{table}

Although our model requires relatively higher FLOPs, its streamlined design yields competitive inference speed. Specifically, CMFPF is only noticeably slower than the four lightweight baselines (MAC\cite{hudson2018compositional}, MCAN\cite{yu2019deep}, RSVQA\cite{lobry2020rsvqa}, DVQA\cite{wen2021debiased}), comparable to SAN\cite{yang2016stacked} and FEH\cite{yuan2022easy}, and substantially faster than SGA\cite{tosato2024segmentation}, EarthVQA\cite{wang2024earthvqa}, MQVQA\cite{zhang2023multi}, and RSIVQA\cite{Zheng2021Mutual}. Overall, our framework emphasizes cross-modal prompting and the modeling of source–tampered region relations, achieving moderate parameter size and acceptable inference speed despite higher computational cost. With further engineering-level optimization, the efficiency of CMFPF could be further improved.

\vspace{-5pt}
\subsection{Transferability experiments}
\noindent \textbf{Baseline Comparison.}  The RS-TQA and RS-TQA-B datasets extend copy-move tampering scenarios by incorporating blurred tampering types, enabling a comprehensive assessment of model robustness and transferability within the RSCMQA task. Experimental results, summarized in Table \ref{tab:tqa}, indicate that the increased dataset size facilitates richer feature learning, leading to generally strong baseline model performance. However, our method consistently demonstrates superior stability and accuracy.
CMFPF outperforms the second-best model on RS-TQA, achieving a 2.79\% improvement in OA and a 3.78\% increase in AA. On RS-TQA-B, CMFPF further enhances OA by 3.40\% and AA by 3.41\%. These results confirm that integrating tampered region prompts effectively enables accurate question answering across diverse tampering scenarios. Our proposed approach exhibits strong transferability, maintaining high performance across multiple tampering types.

\subsection{Ablation Experiments}
\noindent \textbf{Module Selection.}
Although encoder selection is not the primary focus of this study, we explored various feature extraction modules, with experimental results summarized in Table \ref{tab:module_analysis}. Swin Transformer \cite{liu2021swin} and U-Net \cite{ronneberger2015u} were employed to generate tampering mask prompts, with the results indicating that the U-Net module performed better. This may be attributed to the stronger capability of CNNs in extracting local detail features. ResNet-152 \cite{he2016deep} and ViT-B were selected as representatives of CNN-based and Transformer-based visual encoders, respectively, with ViT-B showing a slight overall advantage. LSTM \cite{hochreiter1997long}, and BERT were chosen as representatives of traditional text encoders and Transformer-based text encoders, respectively, with BERT demonstrating a significant advantage in this experiment. The results suggest that changes in the text head caused greater perturbations to the experimental outcomes compared to changes in the visual head. In summary, the combination of U-Net, ViT-B, and BERT consistently achieved superior performance across all three datasets.

\begin{table}[!t]
\centering
\caption{Compared results of multimodal aggregation modules.}
\label{tab:stma}
\resizebox{1\linewidth}{!}{
\begin{tabular}{c|cc|cc|cc}
\midrule[0.8pt]
\multirow{2}{*}{Text Prompt} & \multicolumn{2}{c|}{RS-CMQA} & \multicolumn{2}{c|}{RS-CMQA-B} & \multicolumn{2}{c}{Real-RSCM} \\
 & OA & AA & OA & AA & OA & AA \\ 
\midrule[0.8pt]
Cross-Attention & 89.63 & 87.87 & 68.11 & 68.15 & 86.22 & 84.89 \\
Co-Attention & 90.00 & 88.17 & 69.46 & 69.55 & 86.03 & 84.37 \\
Q-Former & 90.77 & 89.14 & 71.82 & 71.90 & 87.89 & 86.18 \\
AAUE & 91.49 & 89.76 & 71.83 & 71.88 & 89.52 & 88.38 \\
OGA & 91.55 & 89.85 & 72.22 & 72.25 & 90.98 & 89.39 \\ 
\midrule[0.8pt]
\textbf{STMA} & \textbf{93.85} & \textbf{92.44} & \textbf{75.51} & \textbf{75.52} & \textbf{92.86} & \textbf{91.54} \\
\midrule[0.8pt]
\end{tabular}
}
\end{table}

\vspace{5pt}
\noindent \textbf{Multimodal Aggregation Module Comparison.} 
The proposed STMA module is designed to extract information from the source and tampered regions of manipulated images, capturing their differences and similarities to provide cross-modal tampering prompts for the textual modality. To evaluate its effectiveness, we compared STMA with several representative multimodal aggregation methods, including Cross-Attention \cite{vaswani2017attention}, Co-Attention \cite{yu2019deep}, Q-Former \cite{li2023blip}, Adaptive Aggregation of Uni-modal Experts (AAUE) \cite{xu2023managertower}, and Object-Guided Attention (OGA) \cite{wang2024earthvqa}. Cross-Attention represents a classical approach to cross-modal feature aggregation. Co-Attention and Q-Former are widely adopted in the VQA domain for multimodal information fusion. AAUE and OGA are recent developments, introduced in ManagerTower \cite{xu2023managertower} and EarthVQA \cite{wang2024earthvqa}, respectively. Comparative results of these multimodal aggregation modules, presented in Table \ref{tab:stma}, demonstrate that STMA consistently outperforms existing approaches. By explicitly modeling both the disparities and associations between tampered and source regions, STMA provides enhanced feature prompts to the textual modality, thereby significantly improving question-answering accuracy.

\begin{figure}[!t]
    \centering
    \includegraphics[width=1\linewidth]{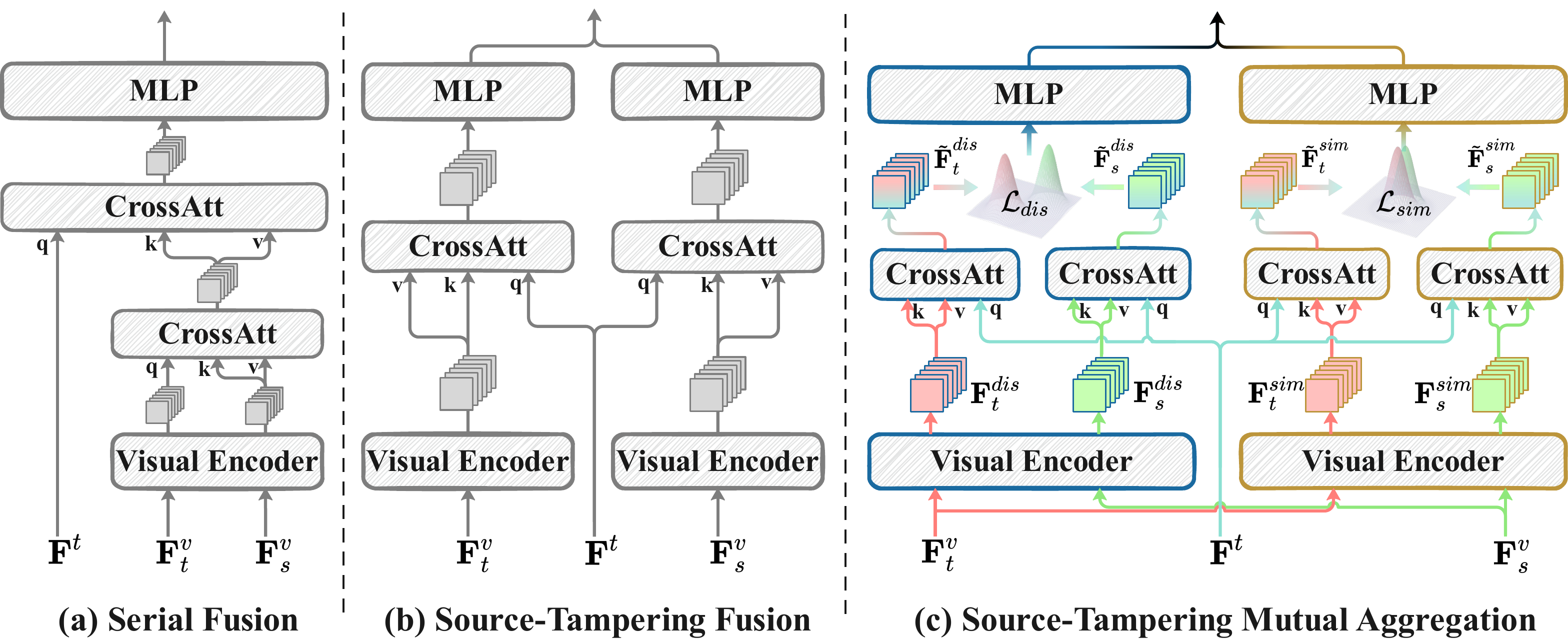}
    \caption{Various multimodal tampering prompt modules used in ablation experiments and the STMA module adopted in CMFPF.}
    \label{fig:stma}
\end{figure}

\noindent \textbf{Ablation Study of STMA Module.} 
The STMA module is the core component of the CMFPF framework, designed to extract both similarities and differences between source and tampered regions in manipulated images. By explicitly modeling these relationships, STMA provides effective cross-modal tampering prompts for the textual modality. As illustrated in Figure~\ref{fig:stma}, we explored two structural variants during the evolution of STMA. The Serial Fusion (SF) strategy sequentially fuses source and tampered features, which leads to premature integration and insufficient cross-modal prompting. In contrast, the Source-Tampering Fusion (STF) strategy associates textual queries with source and tampered features separately, thereby enhancing prompting capability. These explorations highlight the importance of modeling source and tampered regions individually rather than merging them too early.

\begin{table}[!t]
\centering
\caption{Intra-module ablation study of STMA, with experiments conducted on the Real-RSCM dataset.}
\label{tab:stma_ablation}
\resizebox{1\linewidth}{!}{%
  {
  \renewcommand{\arraystretch}{1.0} 
\begin{tabular}{ccccccc}
\midrule[0.8pt]
SF                   & STF                  & D.S. CrossAtt & Sim. Loss & Dis. Loss & OA    & AA    \\ \midrule[0.8pt]
$\checkmark$         &                      &      &      &       & 88.46 & 87.36 \\
                     & $\checkmark$         &      &      &       & 91.96 & 90.79 \\ \midrule[0.8pt]
                     &                      & $\checkmark$    &      &       & 91.38 & 90.22 \\
\multicolumn{1}{l}{} & \multicolumn{1}{l}{} & $\checkmark$    & $\checkmark$    &       & 92.23 & 90.98 \\
                     &                      & $\checkmark$    & $\checkmark$    & $\checkmark$ &\textbf{92.86} & \textbf{91.54} \\ \midrule[0.8pt]
\end{tabular}
  }
}
\end{table}

The backbone of STMA is the Dual-branch Similarity–Difference Cross-Attention (D.S. CrossAtt), consisting of four Cross-Attention layers that establish fine-grained interactions between visual features and textual queries. On this backbone, STMA applies KL-based constraints to two branches: the Similarity Branch, optimized with $\mathcal{L}_{\text{sim}}$ (Eq.~\hyperref[eq:eq12]{12a}), and the Disparity Branch, optimized with $\mathcal{L}_{\text{dis}}$ (Eq.~\hyperref[eq:eq12]{12b}). $\mathcal{L}_{\text{sim}}$ and $\mathcal{L}_{\text{dis}}$ together constitute the overall KL loss $\mathcal{L}_{\text{kl}}$(Eq.~\hyperref[eq:eq12]{12c}). The Similarity Branch encourages alignment when source and tampered regions share semantic consistency, while the Disparity Branch emphasizes differences to highlight tampering cues. By jointly optimizing these two objectives, STMA provides discriminative cross-modal prompts for the textual modality.

Ablation experiments demonstrate the necessity of these constraints, as presented in Table \ref{tab:stma_ablation}. Using only the D.S. CrossAtt backbone without $\mathcal{L}_{\text{sim}}$ and $\mathcal{L}_{\text{dis}}$ weakens similarity–difference modeling, resulting in inferior prompting ability. Disabling either branch individually also degrades performance, confirming their complementary roles. The full STMA, with D.S. CrossAtt and both KL losses enabled, achieves the best accuracy, showing that similarity and disparity modeling are jointly essential for robust tampering-aware multimodal aggregation.

\vspace{5pt}
\noindent \textbf{Comparison of Visual Tampering Prompts.}
STMA has demonstrated strong effectiveness in providing tampering prompts for the textual modality. To further investigate its applicability, we explored its use in the visual modality by applying the same tampering prompt strategy used in the textual modality. Specifically, we extracted relational and differential information from tampered regions using STMA and integrated these features into encoded image representations via post-fusion. However, this approach did not yield optimal results. Instead, a pre-fusion strategy—overlaying source region and tampered region masks directly onto the original image, as illustrated in Figure \ref{fig:framework}(a)—proved to be more effective. The experimental results, presented in Table \ref{tab:visual}, indicate that this improvement arises from the inherent nature of source and tampered region masks, which, along with the original image, belong to the same visual modality and do not require additional semantic alignment. The incorporation of tampering prompts through post-fusion introduces unavoidable information loss due to redundant feature processing, potentially impairing model performance. Therefore, directly overlaying masks onto the visual modality provides a more effective mechanism for incorporating tampering prompts.

\begin{table}[!t]
\centering
\caption{Comparison results of visual tampering prompt methods}
\label{tab:visual}
\resizebox{1\linewidth}{!}{
\begin{tabular}{c|cc|cc|cc}
\midrule[0.8pt]
\multirow{2}{*}{Visual Prompt} & \multicolumn{2}{c|}{RS-CMQA} & \multicolumn{2}{c|}{RS-CMQA-B} & \multicolumn{2}{c}{Real-RSCM} \\
 & OA & AA & OA & AA & OA & AA \\ \midrule[0.8pt]
STMA & 93.31 & 92.07 & 74.47 & 74.48 & 92.11 & 90.74 \\
\textbf{Pre-fusion} & \textbf{93.85} & \textbf{92.44} & \textbf{75.51} & \textbf{75.52} & \textbf{92.86} & \textbf{91.54} \\
\midrule[0.8pt]
\end{tabular}
}
\end{table}

\begin{table}[!t]
\centering
\caption{Results of prompt ablation experiments}
\label{tab:prompt}
\resizebox{1\linewidth}{!}{
\begin{tabular}{cc|cc|cc|cc}
\midrule[0.8pt]
\multirow{2}{*}{\begin{tabular}[c]{@{}c@{}}Visual\\ Prompt\end{tabular}} & \multirow{2}{*}{\begin{tabular}[c]{@{}c@{}}Text\\ Prompt\end{tabular}} & \multicolumn{2}{c|}{RS-CMQA} & \multicolumn{2}{c|}{RS-CMQA-B} & \multicolumn{2}{c}{Real-RSCM} \\
 &  & OA & AA & OA & AA & OA & AA \\ 
\midrule[0.8pt]
× & × & 83.96 & 81.91 & 64.23 & 64.31 & 78.76 & 76.10 \\
$\checkmark$ & × & 87.58 & 85.67 & 66.26 & 66.29 & 80.55 & 79.26 \\
× & $\checkmark$ & 91.57 & 89.86 & 75.93 & 73.97 & 91.70 & 90.53 \\
$\checkmark$ & $\checkmark$ & \textbf{93.85} & \textbf{92.44} & \textbf{75.51} & \textbf{75.52} & \textbf{92.86} & \textbf{91.54} \\ 
\midrule[0.8pt]
\end{tabular}
}
\end{table}

\begin{table}[!t]
\centering
\caption{Evaluation of template bias under different input settings.}
\label{tab:bias}
\resizebox{1\linewidth}{!}{
\begin{tabular}{ccccc}
\midrule[0.8pt]
                    & Visual Input & Question Input & OA    & AA    \\ \midrule[0.8pt]
Question-only       & ×    & $\checkmark$              & 48.34 & 46.72 \\
Image-only          & $\checkmark$     & ×   & 38.48 & 32.15 \\
Rewritten Questions & $\checkmark$     & Rewritten      & 90.64 & 89.20 \\ \midrule[0.8pt]
Original            & $\checkmark$     & $\checkmark$              & 92.86 & 91.54 \\ \midrule[0.8pt]
\end{tabular}
}
\end{table}

\vspace{5pt}
\noindent \textbf{Prompts Ablation.}
In CMFPF, the Tampering Detection Branch generates source and tampered region information, which is incorporated into the model via the mask overlay for visual features and the STMA module for the textual features. Ablation experiments were conducted on these prompts, and the results are presented in Table \ref{tab:prompt}, showing that both types of prompts contributed positively. Notably, when using textual and visual prompts separately, textual prompts yielded a greater performance improvement than visual prompts. This aligns with the findings from the Module Selection experiment, which demonstrated that variations in the text head had a more substantial effect on the results. These insights suggest that semantic enhancement of textual features plays a crucial role in improving question-answering accuracy in the RSCMQA task, highlighting an avenue for further research.

\vspace{5pt}
\noindent \textbf{Analysis of Template Bias and Multimodal Dependency.}
To quantitatively evaluate the influence of template bias, we conduct three complementary experiments on the Real-RSCM dataset, and the results are summarized in Table~\ref{tab:bias}. It is worth noting that the questions in Real-RSCM are not directly generated from manually designed templates. Instead, as illustrated in Figure \ref{fig:questions}, each template is first rewritten into multiple semantically equivalent yet linguistically diverse expressions using a large language model (LLM), followed by manual verification to ensure semantic consistency and answer correctness. This LLM-based question generalization pipeline is specifically introduced during dataset construction to alleviate repetitive syntactic patterns while preserving the semantic meaning of the original questions.

Specifically, the Question-only setting removes all visual inputs and predicts answers solely from textual questions, aiming to evaluate whether the model can exploit linguistic priors or repetitive question patterns. Conversely, the Image-only setting removes question inputs to examine whether visual information alone is sufficient for answering. In addition, the Rewritten Questions setting reconstructs the test questions using the same LLM-based question generalization pipeline shown in Figure \ref{fig:questions} while keeping the corresponding images and answers unchanged, thereby evaluating the robustness of the proposed framework to diverse linguistic expressions.

As shown in Table~\ref{tab:bias}, both the Question-only and Image-only settings exhibit substantially lower performance than the complete multimodal model, indicating that neither textual templates nor visual information alone is sufficient for reliable question answering. More importantly, when the original questions are replaced by LLM-generated paraphrases, the proposed framework still achieves 90.64\% OA and 89.20\% AA, with only marginal decreases of 2.15\% and 2.24\%, respectively, compared with the original test set. These results demonstrate that the proposed benchmark effectively alleviates template bias through the LLM-based question generalization process, and that CMFPF primarily relies on multimodal understanding of visual forgery cues rather than memorizing repetitive question templates or syntactic patterns.

\vspace{5pt}
\noindent \textbf{Forgery Detection Loss Ablation.}
Mean Absolute Error (MAE) and Root Mean Square Error (RMSE) are commonly used loss functions for regional regression, corresponding to the L1 and L2 norms in mathematics, respectively. Previous studies have shown that MAE is more robust to outliers, while RMSE is more sensitive to them \cite{willmott2005advantages,chai2014root}. Their performance varies across different tasks, and they are sometimes used in combination. In the RSCMQA task, using only RMSE as the loss function for the Tampering Detection Branch yields the best results, as shown in Table \ref{tab:rmse}.

\begin{table}[!t]
\centering
\caption{Results of forgery detection loss ablation experiments}
\label{tab:rmse}
\resizebox{1\linewidth}{!}{
\begin{tabular}{cc|cc|cc|cc}
\midrule[0.8pt]
\multirow{2}{*}{MAE} & \multirow{2}{*}{RMSE} & \multicolumn{2}{c|}{RS-CMQA} & \multicolumn{2}{c|}{RS-CMQA-B} & \multicolumn{2}{c}{Real-RSCM} \\
 &  & OA & AA & OA & AA & OA & AA \\ 
 \midrule[0.8pt]
$\checkmark$ & × & 90.34 & 88.59 & 70.81 & 70.81 & 89.99 & 88.74 \\
× & $\checkmark$ & \textbf{93.85} & \textbf{92.44} & \textbf{75.51} & \textbf{75.52} & \textbf{92.86} & \textbf{91.54} \\
$\checkmark$ & $\checkmark$ & 93.27 & 91.69 & 74.76 & 74.75 & 92.51 & 91.19 \\ 
\midrule[0.8pt]
\end{tabular}
}
\end{table}

\begin{table}[!t]
\centering
\caption{Ablation study on loss functions, with experiments conducted on the Real-RSCM dataset.}
\label{tab:loss_ablation}
\resizebox{1\linewidth}{!}{
{\large
\begin{tabular}{lccccc}
\midrule[0.8pt]
\textbf{Configuration} & \textbf{CE Loss}& \textbf{RMSE Loss} & \textbf{KL Loss} & \textbf{OA} & \textbf{AA} \\ \midrule[0.8pt]
Full Loss (CE + RMSE + KL)   & $\checkmark$ & $\checkmark$ & $\checkmark$ & \textbf{92.86} & \textbf{91.54}\\
– Tampering Perception Loss  & $\checkmark$ & × & $\checkmark$ & 82.04 & 81.30 \\
– KL Divergence              & $\checkmark$ & $\checkmark$ & × & 91.43 & 90.27 \\
\midrule[0.8pt]
\end{tabular}
}}
\end{table}

\vspace{5pt}
\noindent \textbf{Ablation Study on Loss Functions.}
In addition to the cross-entropy classification loss (CE), which serves as the fundamental objective of the task and cannot be removed, our framework incorporates two auxiliary losses: the root mean square error loss (RMSE) and the KL divergence loss (KL). RMSE is applied in the tampering-aware network to stabilize the prediction of tampered regions, ensuring numerical consistency and spatial accuracy. $\mathcal{L}_{\text{kl}}$(Eq.~\hyperref[eq:eq12]{12c}) is imposed on the similarity and disparity branches of STMA, with $\mathcal{L}_{\text{sim}}$ (Eq.~\hyperref[eq:eq12]{12a}) encouraging alignment when source and tampered regions share semantic consistency, and $\mathcal{L}_{\text{dis}}$ (Eq.~\hyperref[eq:eq12]{12b}) emphasizing differences to highlight tampering cues.

The ablation results in Table~\ref{tab:loss_ablation} demonstrate the complementary roles of these auxiliary losses. Removing RMSE leads to unstable tampering localization and weaker generalization. Excluding KL loss diminishes the ability of STMA to model similarity and disparity, resulting in inferior cross-modal prompting. The full model, trained with CE, RMSE, and KL jointly, achieves the best performance, confirming that the auxiliary losses are essential for robust tampering-aware multimodal aggregation while CE remains the indispensable backbone objective.

\begin{table}[!t]
\centering
\caption{Overall ablation analysis of CMFPF architecture, with experiments conducted on the Real-RSCM dataset.}
\label{tab:cmfpf_ablation}
\resizebox{1\linewidth}{!}{%
  {\Large 
  \renewcommand{\arraystretch}{1.0} 
  \begin{tabular}{ccccccc}
  \midrule[1.3pt]
  \textbf{Text Head} & \textbf{Vis. Head} & \textbf{Tampering Head} & \textbf{Text Prompt} & \textbf{Vis. Prompt} & \textbf{OA} & \textbf{AA} \\ \midrule[1.3pt]
  $\checkmark$ & $\checkmark$ &  &  &  & 71.09 & 70.13 \\
  $\checkmark$ & $\checkmark$ & $\checkmark$ &  &  & 78.31 & 77.06 \\
  $\checkmark$ & $\checkmark$ & $\checkmark$ & $\checkmark$ &  & 91.84 & 90.41 \\
  $\checkmark$ & $\checkmark$ & $\checkmark$ & $\checkmark$ & $\checkmark$ & \textbf{92.86} & \textbf{91.54} \\ 
  \midrule[1.3pt]
  \end{tabular}
  }
}
\end{table}

\vspace{5pt}
\noindent \textbf{Overall Ablation Analysis of CMFPF Architecture.}
To evaluate the contribution of each component in the CMFPF framework, we conducted an ablation study as shown in Table~\ref{tab:cmfpf_ablation}. The architecture includes five modules: the Text Head and Visual Head for modality-specific encoding, the Tampering Head for manipulation-aware feature extraction, and two prompting mechanisms—Text Prompt and Visual Prompt—for cross-modal guidance.

With only the Text and Visual Heads, performance remains limited, indicating that basic multimodal encoding is insufficient for tampering-aware reasoning. Adding the Tampering Head improves results by capturing manipulation cues. Incorporating the Text Prompt then leads to a substantial gain, as the textual modality benefits from explicit tampering-aware guidance; in particular, the Text Prompt significantly enhances cross-modal alignment. Finally, enabling both Text and Visual Prompts yields the best performance, validating their complementary roles and confirming that full integration is essential for robust tampering-aware multimodal reasoning.

\begin{figure}[!t]
    \centering
    \includegraphics[width=1\linewidth]{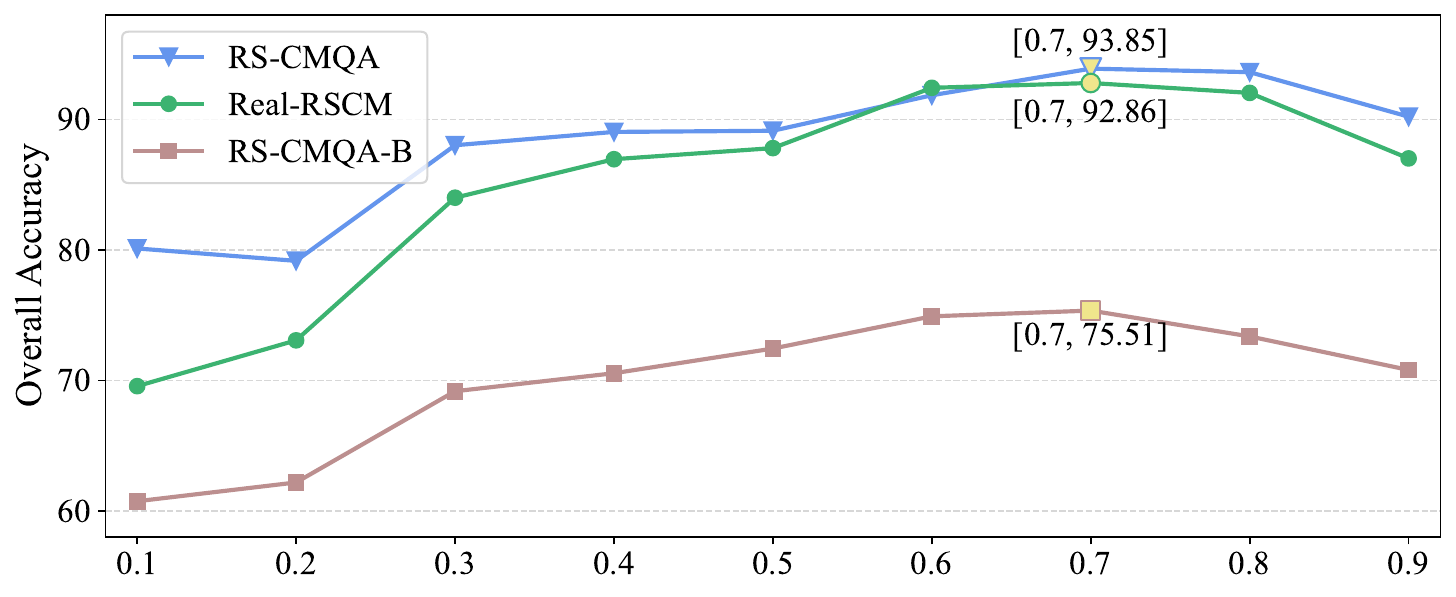}
    \caption{Results with varied hyperparameter \(\alpha\).
    }
    \label{fig:hyper}
\end{figure}

\vspace{5pt}
\noindent \textbf{Hyperparameter settings.} In the loss function, \(\alpha\) serves as a hyperparameter to balance the forgery detection loss and the VQA loss. To determine an appropriate value, we conducted a grid search with a step size of 0.1 within the range of 0.1 to 0.9, as illustrated in Figure \ref{fig:hyper}. Experimental results indicate that the range of \(\alpha\) between 0.6 and 0.8 is relatively optimal, with CMFPF achieving the best performance across all three datasets when \(\alpha\) is set to 0.7.

\subsection{Comparison with Traditional CMFD Methods}

\begin{table}[!t]
\centering
\caption{Performance comparison of copy–move forgery detection methods on the Real-RSCM dataset}
\begin{tabular}{cccccc}
\midrule[0.8pt]
Methods & Precision & Recall & F1  & OA & AA \\
\midrule[0.8pt]
U-Net\cite{ronneberger2015u} & 0.3432 & 0.1118  & 0.1687 & - & - \\
Dense-Field\cite{cozzolino2015efficient}  & 0.7465 & 0.8128 & 0.7782 & - & - \\
BusterNet\cite{wu2018busternet} & 0.8594 & 0.8390 & 0.8491 & - & -\\
DOA-GAN\cite{islam2020doa}   & 0.6783 & 0.5944 & 0.6336  & - & - \\
CMCF-Net\cite{xiong2023cmcf}   & 0.9435 & 0.8595 & 0.8996  & - & - \\
UCM-Net\cite{weng2023ucm}  &  0.8852 & 0.9220 & 0.9032  & - & -\\
IMNet\cite{wang2024object}  & 0.9666 & 0.9233 & 0.9445 & - & - \\
ST-Net\cite{10856558}  & 0.9487 & 0.9113 & 0.9296  & - & -\\
\midrule[0.8pt]
\textbf{CMFPF} & 0.9616 & 0.9136 & 0.9370  & 92.86 & 91.54\\
\midrule[0.8pt]
\end{tabular}
\label{tab:cmfd}
\end{table}

Traditional Copy-Move Forgery Detection (CMFD) methods mainly focus on localizing tampered regions and identifying source areas, typically producing binary masks without forgery-oriented multimodal reasoning or multimodal integration. In our framework, the CMFP module fulfills this role by providing tampering-aware cues, though it is not specifically optimized for CMFD. To ensure fair comparison, we evaluate our CMFPF module against representative CMFD approaches, including traditional methods and deep-neural-network-based detection methods. For evaluation, we adopt the standard metrics Precision, Recall, and F1-score, defined as:
\begin{equation}
\text{Precision} = \frac{\text{TP}}{\text{TP} + \text{FP}},
\end{equation}
\begin{equation}
\text{Recall} = \frac{\text{TP}}{\text{TP} + \text{FN}},
\end{equation}
\begin{equation}
\text{F1} = \frac{2 \cdot \text{Precision} \cdot \text{Recall}}{\text{Precision} + \text{Recall}},
\end{equation}
where $TP$, $FP$, and $FN$ denote true positives, false positives, and false negatives, respectively.

As presented in Table~\ref{tab:cmfd}, the experimental results show that using CMFPF to accomplish CMFD tasks consistently outperforms the majority of traditional CMFD methods, including classical dense-field approaches such as Dense-Field\cite{cozzolino2015efficient}, and significantly surpasses the performance ceiling of employing U-Net\cite{ronneberger2015u} or its enhanced variant UCM-Net\cite{weng2023ucm} alone. Moreover, CMFPF achieves superior accuracy compared with several representative deep learning architectures, such as the end-to-end BusterNet\cite{wu2018busternet}, the adversarial learning based DOA-GAN\cite{islam2020doa}, the multiscale context fusion enhanced CMCF-Net\cite{xiong2023cmcf}, and the lightweight high-efficiency ST-Net\cite{10856558}. Although CMFPF is not specifically optimized for CMFD, the multi-task loss joint optimization strategy enables it to achieve highly competitive detection accuracy. In fact, CMFPF only falls slightly behind the most advanced state-of-the-art CMFD architectures, such as IMNet\cite{wang2024object}, which leverages inconsistency mining for refined detection, while clearly outperforming classical baselines. Such competitiveness validates the effectiveness of our multi-task optimization strategy, which has been shown to improve generalization and robustness by leveraging shared representations across tasks~\cite{ruder2017overview,li2022blip}. Moreover, unlike traditional CMFD methods that merely output binary masks or localized regions, our framework integrates tampering detection with forgery-oriented multimodal reasoning, ensuring that the tampering-aware cues provided by CMFPF are sufficiently reliable to support the overall multimodal QA evaluation. In future work, we plan to incorporate more advanced tampering detection architectures to further enhance localization under challenging remote sensing scenarios.

\subsection{Comparison with multimodal large language models}

\begin{table}[!t]
\centering
\caption{Zero-shot comparison with representative instruction-tuned MLLMs on the RS-CMQA dataset.}
\begin{tabular}{lccc}
\midrule[0.8pt]
Method       & Params & OA    & AA    \\ \midrule[0.8pt]
LLaVA-1.5\cite{liu2024improved} & 7B     & 45.96 & 43.55 \\
InstructBLIP\cite{dai2023instructblip} & 7B     & 42.22 & 39.98 \\
InternVL-3.5\cite{wang2025internvl3}   & 14B    & 49.99 & 47.84 \\
Qwen3-VL\cite{bai2025qwen3}       & 32B    & 51.60 & 49.98 \\ 
\midrule[0.8pt]
\textbf{CMFPF}      & 54M    & 92.86 & 91.54 \\ \midrule[0.8pt]
\end{tabular}
\label{tab:mllm}
\end{table}

Recent instruction-tuned multimodal large language models (MLLMs) have achieved remarkable performance on general visual reasoning and VQA tasks. To further contextualize the difficulty of the proposed RS-CMQA dataset, we additionally benchmark four representative open-source MLLMs, including LLaVA-1.5~\cite{liu2024improved}, InstructBLIP~\cite{dai2023instructblip}, InternVL-3.5~\cite{wang2025internvl3}, and Qwen3-VL~\cite{bai2025qwen3}.
All models are evaluated in a zero-shot setting without task-specific fine-tuning. Each model receives only the original remote sensing image and the corresponding question. The purpose of this experiment is not to compare fully task-adapted models under identical training protocols, but to assess the out-of-the-box capability of modern MLLMs on the proposed benchmark.

The results are summarized in Table~\ref{tab:mllm}. Qwen3-VL achieves the best performance among the evaluated MLLMs with an OA of 51.60\% and an AA of 49.98\%, while the remaining models achieve OA and AA around 42--50\%. Despite their strong general reasoning capability and significantly larger parameter scales (7B--32B), all evaluated MLLMs perform substantially worse than CMFPF, which achieves 92.86\% OA and 91.54\% AA with only 54M parameters.

This performance gap stems from the unique challenges of remote sensing copy-move forgery understanding. Unlike conventional VQA, the task requires accurate localization of both the source and tampered regions, reliable correspondence modeling, and fine-grained spatial reasoning. Generic MLLMs mainly rely on holistic visual-language understanding and lack explicit mechanisms for exploiting forgery-related localization cues. In contrast, CMFPF transforms the predicted source and tampered regions into visual prompts through the proposed STMA module, enabling localization-guided multimodal reasoning. These results demonstrate both the challenging nature of the proposed dataset and the necessity of explicit localization guidance for copy-move forgery question answering.

Nevertheless, these results do not imply limitations of modern MLLMs. Instead, they highlight that generic multimodal reasoning alone is insufficient for this task. We believe that integrating explicit forgery localization cues into powerful MLLMs through localization-guided visual prompting is a promising future direction, combining the strong reasoning capability of MLLMs with specialized forgery perception for more robust remote sensing forgery analysis.

\subsection{Examples and Visualizations}
\begin{figure}[!t]
    \centering
    \label{fig:final}
    \includegraphics[width=1\linewidth]{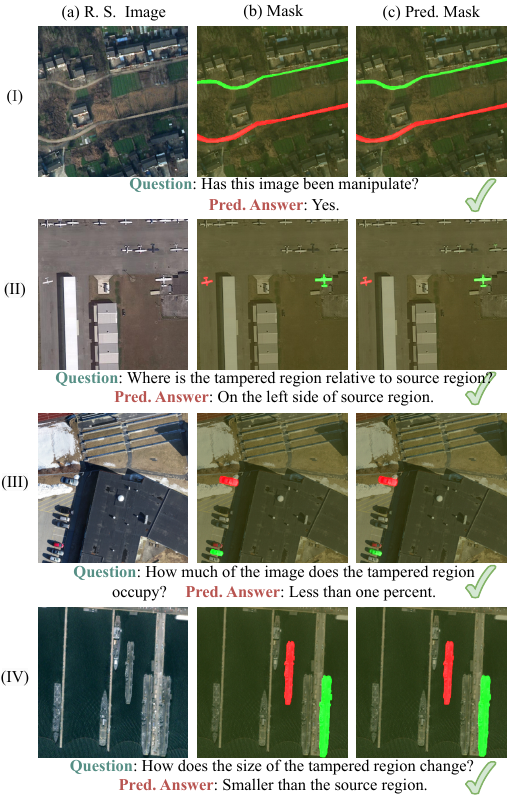}
    \caption{Question-Answering Examples of CMFPF on the RSCMQA Task. (a) Input remote sensing image, (b) Ground truth masks for source and tampered regions, (c) Predicted masks for source and tampered regions.
    }
    \vspace{-10pt}
    \label{fig:final10}
\end{figure}

As demonstrated in previous experiments, the proposed CMFPF framework achieves strong performance in the RSCMQA task, providing accurate answers to the majority of questions. Figure \ref{fig:final10} presents several question-answering examples, each displaying the original image, ground truth masks for source and tampered regions, and predicted masks for these regions. In these visualizations, green represents source regions, while red indicates tampered areas. It is important to note that both ground truth and predicted masks are inherently binary; the use of red and green overlays is solely for visualization purposes. There are four representative cases—case \hyperref[fig:final]{I} ambiguous boundaries; case \hyperref[fig:final]{II} numerous visually similar objects; case \hyperref[fig:final]{III} small-object tampering; and case \hyperref[fig:final]{IV} large-object tampering. In these scenarios, CMFPF exhibits near-perfect performance, accurately delineating source and tampered regions and correctly answering the corresponding questions based on tampering prompts.

\begin{figure}[!t]
    \centering
    \label{fig:badcase}
    \includegraphics[width=0.98\linewidth]{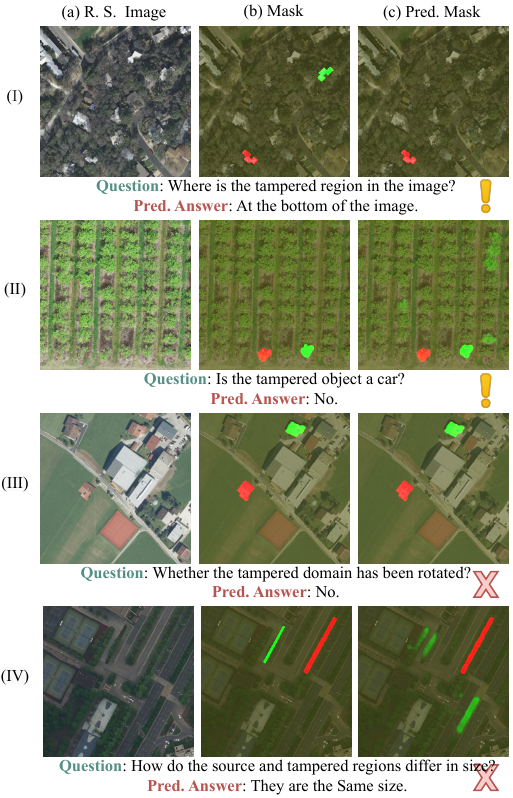}
    \caption{Negative Examples of CMFPF on the RSCMQA Task. (a) Input remote sensing image, (b) Ground truth masks for source and tampered regions, (c) Predicted masks for source and tampered regions.
    }
    \vspace{-10pt}
    \label{fig:badcase10}
\end{figure}

In complex scenarios, especially when scaling and rotation transformations are involved, our model still faces certain challenges. As shown in Figure \ref{fig:badcase10}, we present several representative failure cases. In Example \hyperref[fig:badcase]{I}, the model only identified the tampered region but failed to locate the source region. This may be because the question was posed as an independent query about the tampered area, leading the model to overlook the source region. Moreover, the tampering in this image was subtle and involved rotation, making it difficult even for human observers to distinguish. In Example \hyperref[fig:badcase]{II}, the model exhibited slight misactivation in the source region, possibly due to the presence of many visually similar objects, which became even harder to distinguish after rotation. Despite imperfect region discrimination, CMFPF still provided correct answers in both cases. In Example \hyperref[fig:badcase]{III}, the model accurately identified both the source and tampered regions, but misjudged whether the tampered object had been rotated, indicating that the model’s conceptual understanding of rotation remains insufficient. This suggests that additional rotation-detection mechanisms may be required. In Example \hyperref[fig:badcase]{IV}, the tampered building underwent scaling, and multiple similar shapes of different sizes coexisted in the same area. This posed a severe challenge, leading the model to misactivate several possible regions and ultimately produce an incorrect answer.

Rotation and scale variations represent the most challenging cases in copy-move forgery perception. These geometric transformations distort spatial correspondence and increase ambiguity among visually similar objects, often resulting in false activations or misjudgments. To overcome these challenges, future work may incorporate rotation-invariant and scale-invariant mechanisms, such as transformation-equivariant feature extraction, multi-scale attention modules, or geometric consistency constraints. Embedding these modules into the CMFPF framework could enhance robustness in complex tampering scenarios and improve accuracy under severe geometric variations.

It should be noted that these failure cases were deliberately selected to comprehensively demonstrate the model’s performance. In most situations, the model successfully identifies both source and tampered regions with high accuracy and provides correct answers across different types of questions.

\begin{figure*}[!t]
    \centering
    \includegraphics[width=1\linewidth]{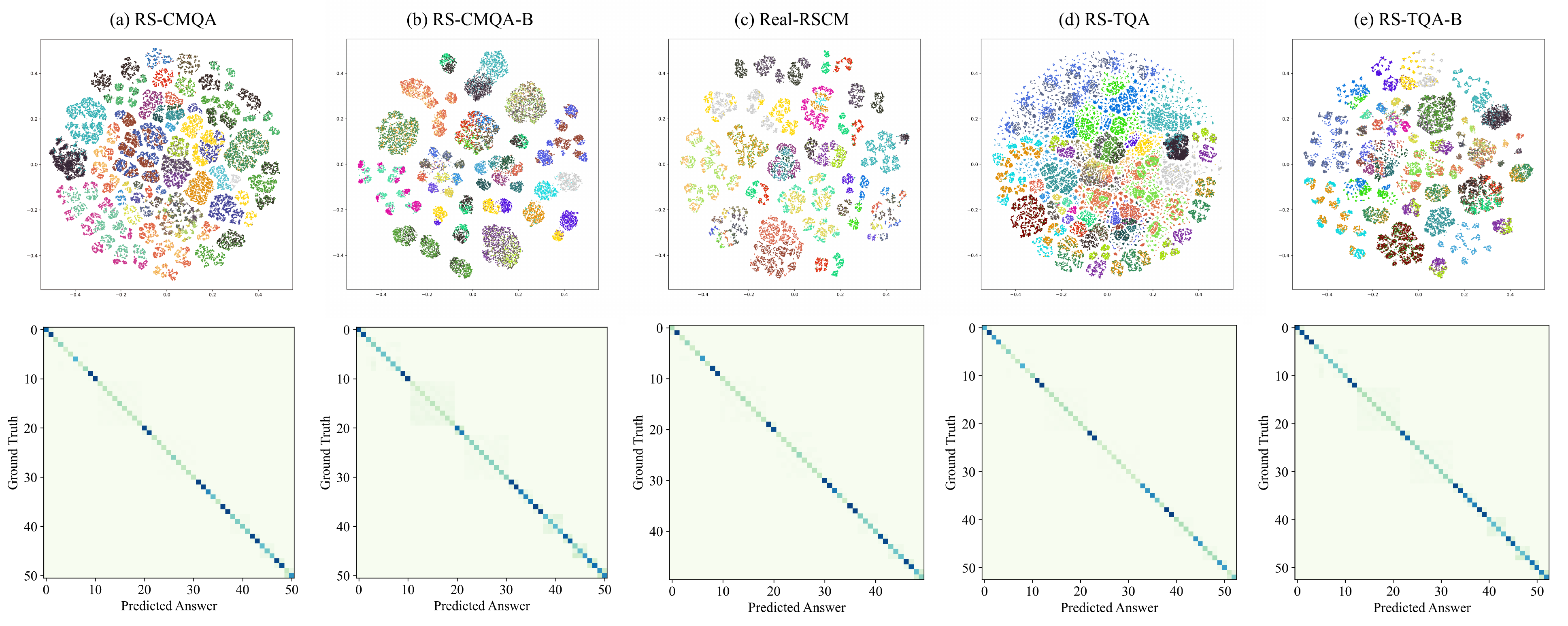}
    \caption{First row: t-SNE-based dimensionality reduction of feature vectors extracted from CMFPF predictions. Second row: Confusion matrices illustrating CMFPF prediction results across five datasets.
    }
    \label{fig:matrix}
\end{figure*}

The feature vectors extracted before the final fully connected layer typically encapsulate rich and comprehensive characteristics, providing insights into the model’s ability to distinguish features. Figure \ref{fig:matrix} presents a t-SNE-based dimensionality reduction visualization of feature vectors predicted by CMFPF across five datasets. The results indicate that CMFPF exhibits strong feature discrimination capability in RS-CMQA, RS-CMQA-B, Real-RSCM, and RS-TQA-B datasets, with clear inter-class separability and clear separation among answer categories. For RS-TQA, although the feature visualization appears relatively sparse, this may be attributed to the large sample size and high complexity of tampering scenarios within the dataset, leading the model to excessively refine feature distinctions. 
This aligns with the expected performance of models with strong discriminative capabilities in complex environments. Although the feature visualization does not exhibit high spatial density, fine-grained feature clusters remain distinct and well-defined, with clear separability between different answer categories. Experimental results confirm that the final feature vector yields highly accurate predictions without any adverse effects.

Figure \ref{fig:matrix} also presents the confusion matrices of CMFPF predictions across five datasets, revealing that most classification results are aligned along the diagonal. No misclassifications in question categories are observed, with prediction errors primarily concentrated in a few challenging cases, such as source region localization and source–tampered region correlation. Overall, CMFPF demonstrates strong performance across all five datasets.

\subsection{Analysis of Localization Error Propagation}

\begin{figure}[!t]
    \centering
    \label{fig:badcase}
    \includegraphics[width=0.98\linewidth]{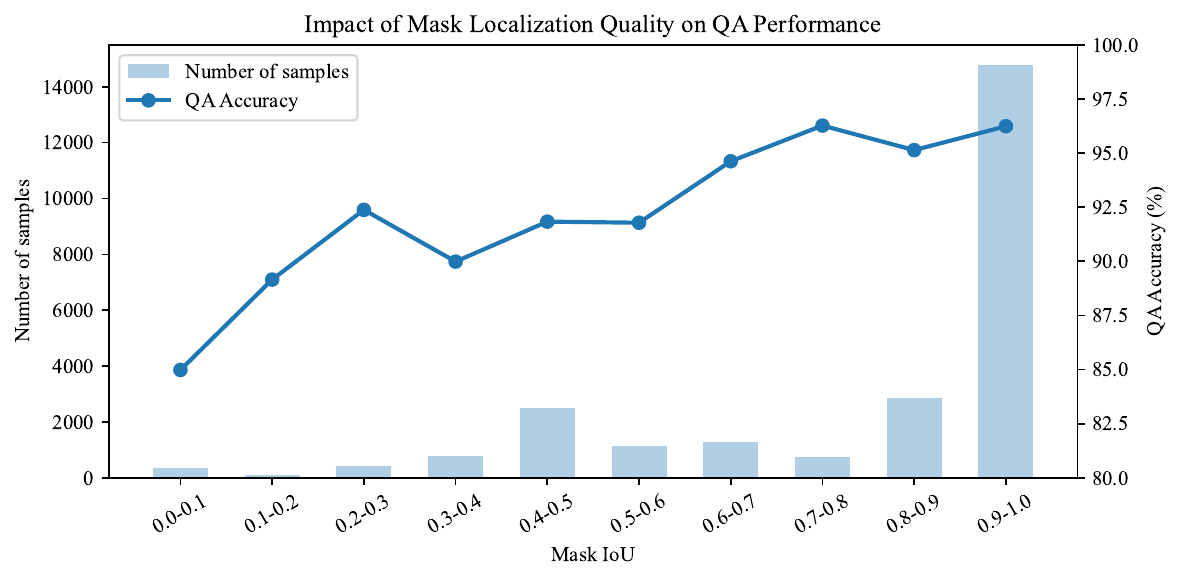}
    \caption{
    Quantitative analysis of the relationship between tampering localization quality and visual question answering performance. 
    The bars represent the number of test samples under different mask IoU intervals, and the curve indicates the corresponding QA accuracy.
    }
    \label{fig:iou}
\end{figure}

Although CMFPF adopts a cascaded architecture, where the tampering detection branch provides visual prompts for the downstream VQA module, localization errors do not necessarily lead to incorrect answers. This is because the predicted source and tampered regions are transformed into soft visual prompts rather than being directly treated as the sole input for answer prediction. These prompts are further integrated with the original visual representation through the proposed STMA module, enabling the reasoning process to exploit complementary semantic information beyond localized regions.

As illustrated in Figure \ref{fig:badcase10}(I)-(II), inaccurate mask boundaries or incomplete localization do not always degrade the final answers, since global visual information and textual reasoning can compensate for moderate localization errors. Conversely, Figure \ref{fig:badcase10}(IV) demonstrates that severe localization errors, especially under large geometric transformations, may introduce misleading visual prompts and consequently affect downstream reasoning. Moreover, Figure \ref{fig:badcase10}(III) indicates that QA failures may also originate from forgery-oriented multimodal reasoning limitations even when localization results are relatively accurate.

To quantitatively evaluate the magnitude of error propagation from tampering localization to downstream QA, we further investigate the relationship between localization quality and QA performance on the Real-RSCM test set. Specifically, we calculate the IoU between the predicted and ground-truth tampering masks as a metric of localization quality. Since CMFPF utilizes both source and target regions to construct visual prompts, the source and target masks are merged through a logical OR operation to represent the complete manipulated region. The test samples are then grouped according to mask IoU intervals with a step size of 0.1, and the corresponding QA accuracy is calculated for each interval.

As shown in Figure \ref{fig:iou}, QA performance exhibits an overall increasing trend as the localization quality improves. When the mask IoU is lower than 0.1, the QA accuracy is 84.97\%, whereas samples with IoU values higher than 0.9 achieve an accuracy of 96.25\%. This demonstrates that localization errors can propagate to downstream reasoning, resulting in an accuracy degradation of 11.28 percentage points under severe localization failures. Nevertheless, the degradation trend remains gradual rather than catastrophic. Even under extremely poor localization conditions (IoU < 0.1), CMFPF still achieves a QA accuracy of 84.97\%, demonstrating its robustness against imperfect localization. This robustness can be attributed to two aspects of the proposed design. First, the prompt construction strategy retains the original image representation together with localized region representations, preventing the reasoning module from relying solely on imperfect masks. Second, the similarity-disparity learning mechanism in STMA encourages the model to capture consistent and discriminative forgery-related cues through cross-modal interaction, which helps reduce the influence of noisy visual prompts.

\section{Conclusion}
\label{sec:6}
In this study, we integrate tampering detection into Remote Sensing Visual Question Answering by introducing a novel task, Remote Sensing Copy-Move Question Answering. To support this task, we have constructed five unique datasets that bridge a critical gap in the field while ensuring comprehensive, balanced, challenging, and generalizable evaluations. Extensive experiments conducted on these datasets establish a robust benchmark for future research. Additionally, we propose the Copy-Move Forgery Perception Framework that injects tampering cues into both textual and visual modalities to guide the model in accurately answering tampering-related questions. Our extensive experimental results demonstrate the superior performance of CMFPF compared to existing models. In future work, we plan to further enrich the datasets by incorporating additional types of image tampering and diversifying the question types. Moreover, we will explore the incorporation of tampered region information into large-scale multimodal models to investigate the reasoning relationship between tampering cues and question answering, thereby advancing the practical application of remote sensing image tampering perception in real-world scenarios.

\bibliographystyle{ieeetr}
\bibliography{bare}
\vspace{-10 mm}

\begin{IEEEbiography}[{\includegraphics[width=1in,height=1.25in,clip,keepaspectratio]{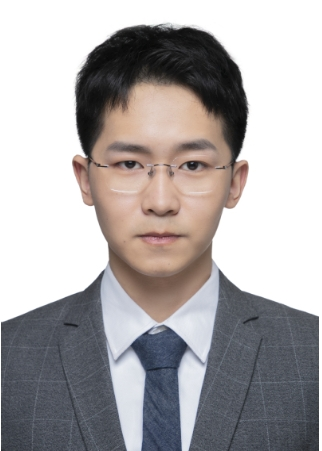}}]{Ze Zhang}
Ze Zhang is currently a master's student at the Faculty of Information Science and Engineering, Ocean University of China, located in Qingdao, China. His research focuses on remote sensing image interpretation and applying multimodal approaches to solve scientific problems. His work aims to integrate diverse data sources to enhance the understanding of complex visual and scientific information.
\end{IEEEbiography}
\vspace{-12 mm}

\begin{IEEEbiography}[{\includegraphics[width=1in,height=1.25in,clip,keepaspectratio]{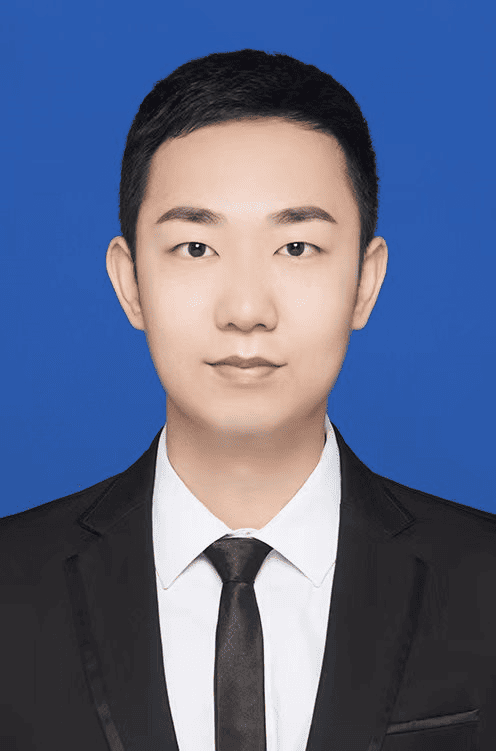}}]{Enyuan Zhao}
received a B.S. in electronics and information engineering from Northeast Agricultural University, China, and an M.S. in computer science and technology from Harbin Engineering University, China, in 2021. He received a Ph.D. in computer science from Ocean University of China, Qingdao, China, which he obtained in 2025. Currently, he works at Hangzhou Institute for Advanced Study, University of Chinese Academy of Sciences, Hangzhou, China. His research interests include deep learning and intelligent multimodal systems.
\end{IEEEbiography}
\vspace{-12 mm}

\begin{IEEEbiography}[{\includegraphics[width=1in,height=1.25in,clip,keepaspectratio]{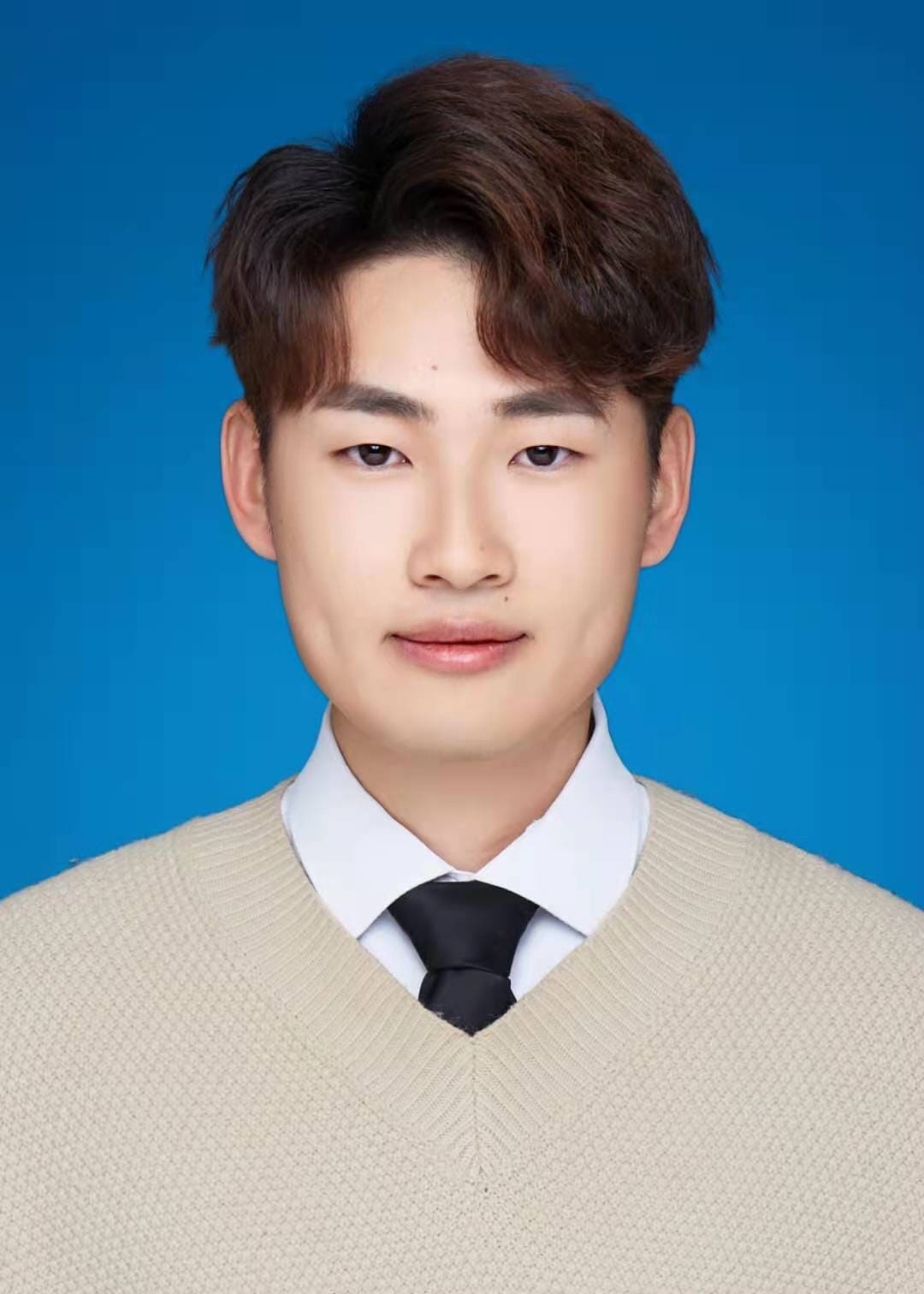}}]{Di Niu}
is currently pursuing his Ph.D. degree at the College of Information Science and Engineering, Ocean University of China.  He received his B.S. degree from the School of Science, Qingdao University of Technology in 2016, and his M.S. degree from the College of Information Science and Engineering, Ocean University of China in 2021.  His research interests include visual question answering, time-series anomaly detection and novel class discovery in the field of fault diagnosis.
\end{IEEEbiography}
\vspace{-12 mm}

\begin{IEEEbiography}[{\includegraphics[width=1in,height=1.25in,clip,keepaspectratio]{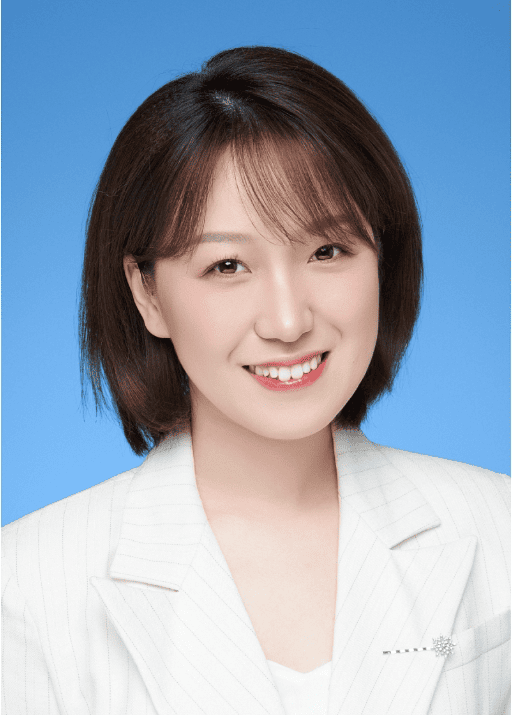}}]{Jie Nie}
(Member, IEEE) received a Ph.D. in Computer Science from Ocean University of China, Qingdao, China, which she obtained in 2011. She furthered her academic pursuits as a Visiting Scholar at the School of Electrical Engineering, University of Pittsburgh, Pennsylvania, USA, from September 2009 to September 2010. Following her doctorate, she conducted post-doctoral research at Tsinghua University, Beijing, China, from 2015 to 2017. Her research focuses on social media and multimedia content analysis.
\end{IEEEbiography}
\vspace{-12 mm}

\begin{IEEEbiography}[{\includegraphics[width=1in,height=1.25in,clip,keepaspectratio]{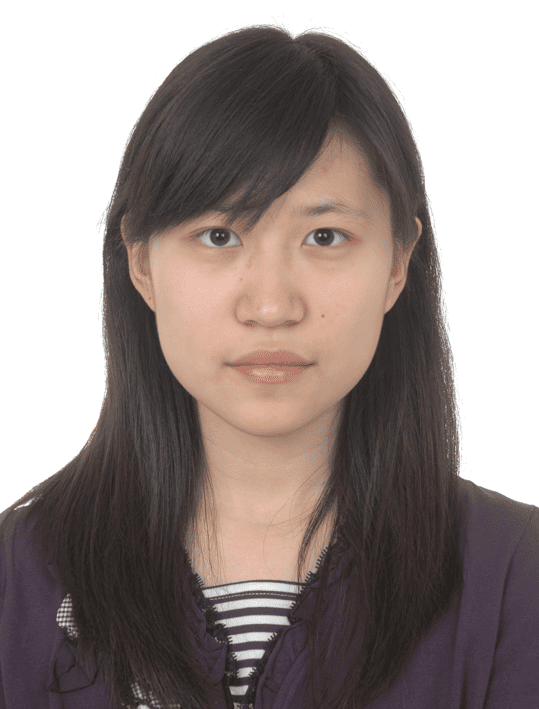}}]{Xinyue Liang}
(Member, IEEE) received her B.S. degree in Communication Engineering from Beijing Jiaotong University in 2014. From 2014 to 2021, she pursued her Master's and Ph.D. degrees at the Royal Institute of Technology in Sweden. She is currently an Associate Professor at Ocean University of China. Her research interests include distributed machine learning, marine big data analytics, and distributed signal analysis.
\end{IEEEbiography}
\vspace{-12 mm}

\begin{IEEEbiography}[{\includegraphics[width=1in,height=1.25in,clip,keepaspectratio]{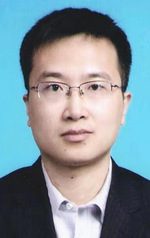}}]{Lei Huang}
(Member, IEEE) is a professor at Ocean University of China. He obtained his Ph.D. from the Institute of Computing Technology, Chinese Academy of Sciences in 2013. His current research focuses on multimedia content analysis and retrieval, machine learning, and intelligent information processing.
\end{IEEEbiography}

\end{document}